\documentclass{siamonline1116}

\newcommand{\be}{\begin{equation}}
\newcommand{\ee}{\end{equation}}
\newcommand{\bc}{\begin{center}}
\newcommand{\ec}{\end{center}}
\newcommand{\bd}{\begin{description}}
\newcommand{\ed}{\end{description}}
\newcommand{\bi}{\begin{itemize}}
\newcommand{\ei}{\end{itemize}}
\newcommand{\pa}{\partial}
\newcommand{\bs}{\boldsymbol}
\newcommand{\bsl}{\bs{\lambda}}
\newcommand{\bx}{\bs{x}}
\newcommand{\bu}{\bs{u}}
\newcommand{\bt}{\bs{\theta}}
\def\RR{ \mathbb R}

\newcommand{\refeq}[1]{Equation \ref{#1}}

\newcommand{\tm}{\textrm}
\newcommand{\tit}{\textit}

\newcommand{\tbf}{\textbf}
\newcommand{\bmat}{\begin{pmatrix}}
\newcommand{\emat}{\end{pmatrix}}
\newcommand{\bsmat}{\left(\begin{smallmatrix}}
\newcommand{\esmat}{\end{smallmatrix}\right)}
\newcommand{\bes}{\begin{equation}\begin{split}}
\newcommand{\ees}{\end{split}\end{equation}}

\newcommand{\rpsk}[1]{\textcolor{black}{PSK: #1}}
\newcommand{\revone}[1]{\textcolor{black}{#1}}
\newcommand{\revtwo}[1]{\textcolor{black}{#1}}

\usepackage{lipsum}
\usepackage{amsfonts}
\usepackage{graphicx}
\usepackage{epstopdf}
\usepackage{algorithmic}
\usepackage{bbold}
\usepackage{caption}
\usepackage{subcaption}
\usepackage{diagbox}
\usepackage{multicol}
\usepackage{makecell}

\ifpdf
  \DeclareGraphicsExtensions{.eps,.pdf,.png,.jpg}
\else
  \DeclareGraphicsExtensions{.eps}
\fi

\numberwithin{theorem}{section}

\newcommand{\TheTitle}{Bayesian model and dimension reduction for uncertainty propagation: applications in random media}

\newcommand{\TheAuthors}{C. Grigo, and P.-S. Koutsourelakis}

\headers{\TheTitle}{\TheAuthors}

\title{{\TheTitle}\thanks{Submitted to the editors \today.
}}

\author{
  Constantin Grigo\thanks{Department of Mechanical Engineering, Technical University of Munich, Germany
    (\email{constantin.grigo@tum.de}, \url{https://www.contmech.mw.tum.de/index.php?id=5}).}
  \and
  Phaedon-Stelios Koutsourelakis\thanks{Department of Mechanical Engineering, Technical University of Munich, Germany (\email{p.s.koutsourelakis@tum.de}.}
}

\usepackage{amsopn}

\usepackage{tikz}
\usetikzlibrary{calc,fit,arrows,arrows.meta, backgrounds}
\pgfdeclarelayer{background}
\pgfdeclarelayer{foreground}
\pgfsetlayers{background,foreground}

\ifpdf
\hypersetup{
  pdftitle={\TheTitle},
  pdfauthor={\TheAuthors}
}
\fi

\begin{document}

\maketitle

\begin{abstract}
Well-established  methods  for  the  solution  of  stochastic  partial  differential  equations  (SPDEs) typically struggle in problems with high-dimensional inputs/outputs.  Such difficulties are only amplified in large-scale applications where even a few tens of full-order model runs are \revtwo{impracticable}.  While dimensionality reduction can alleviate some of these issues, it is not known which and how many features of the (high-dimensional) input are actually predictive of the (high-dimensional) output.  In this paper, we advocate a Bayesian formulation that is capable of performing simultaneous dimension and model-order reduction.  It consists of a component that encodes the high-dimensional input into a low-dimensional set of feature functions by employing sparsity-inducing priors and a decoding component that makes use of the solution of a coarse-grained model in order to reconstruct that of the full-order model. Both components are represented with latent variables in a probabilistic graphical model and are simultaneously trained using Stochastic Variational Inference methods.  The model is capable of quantifying the predictive uncertainty due to the information loss that unavoidably takes place in any model-order/dimension reduction as well as the uncertainty arising from finite-sized training datasets. We  demonstrate  its  capabilities  in  the  context  of  random  media  where  fine-scale  fluctuations  can  give  rise to random inputs with tens of thousands of variables.  With a few tens of full-order model simulations,  the proposed model is capable of identifying salient physical features and produce sharp predictions under different boundary conditions of the full output which itself consists of thousands of components.

\end{abstract}



\section{Introduction}
\label{sec:intro}
%
%
\revtwo{One of the  most difficult obstacles in the application of  uncertainty quantification methods in large-scale engineering problems pertains to the  poor scalability of uncertainty propagation tools in high dimensions.}
The golden standard for such problems i.e. Monte Carlo, exhibits convergence rates that are independent of the dimension of the random input (and output). Nevertheless, for computationally intensive models for which only 10 or 100 runs can be practicably performed, it is of paramount importance to decrease as much as possible the number of simulations needed. This can only be achieved if one can  extract sufficient knowledge from the few simulations  that can be carried out in order to infer the quantities of interest  \cite{rasmussen_bayesian_2003}.

One obvious strategy in overcoming these limitations is the use of surrogates or emulators that are trained on a limited number of runs 
 and can  subsequently substitute the forward model. Amongst existing methods for uncertainty propagation, those based on (generalized) polynomial chaos expansions (gPC, \cite{wiener_homogeneous_1938})  have grown into prominence in recent years with the development of non-intrusive, stochastic collocation approaches \cite{xiu_high_2005,ma_adaptive_2009}. More recent efforts have employed Gaussian Processes (GPs, \cite{bilionis_multi-output_2012,bilionis_multi-output_2013}) or multivariate regression schemes \cite{bilionis_multidimensional_2012}. While all these tools are highly expressive and can potentially  approximate sufficiently well the sought input-output map, they exhibit significant limitations in high input dimensions (e.g. in the hundreds), an instantiation of the well-documented curse of dimensionality \cite{constantine_active_2015}.   One could argue that employing larger, more flexible emulators, e.g. as those arising in the context of Deep Neural Networks \cite{bengio_deep_2015,lecun_deep_2015}, could overcome such problems. We emphasize though that uncertainty propagation problems in computational physics and engineering are not Big Data problems \cite{koutsourelakis_special_2016} and minimizing the number of training data generated by running the full-order simulator is the primary objective.

 A more recent trend to the problem has been based on the use of less-expensive, lower-fidelity models in order to provide accurate estimates of the higher-fidelity quantities of interest \cite{kennedy_predicting_2000}. When combined with statistical learning procedures, such formulations can also yield quantitative estimates of the confidence in the predictions produced \cite{koutsourelakis_accurate_2009}. One of the strengths of such tools stems from the use of lower-fidelity models that retain some of the underlying physics and as such produce outputs that are strongly correlated/dependent with the high-fidelity ones \cite{Perdikaris2015}. The  systematic construction of such lower-fidelity or, more generally, reduced-order models, has also received a lot of attention.  A prominent role in these efforts, at least in the context of PDE-based models, is held by reduced-basis techniques \cite{noor_reduced_1980,quarteroni_reduced_2016,hesthaven_certified_2016} which are based on the identification of a low-dimensional linear subspace in the  solution vector space on which a  Galerkin projection of the governing equations is attempted \cite{veroy_certified_2005,grepl_efficient_2007,maday_generalized_2013,cui_data-driven_2015}. Naturally such an assumption ceases to hold as higher-dimensional inputs are considered and various strategies have been  adopted 
  to address this limitation \cite{elman_reduced_2013,chen_reduced_2017}.

  The potential of dimensionality-reduction methods in overcoming the curse of dimensionality has also been demonstrated by employing  data-driven, nonlinear, manifold learning techniques (e.g. \cite{tenenbaum_global_2000,roweis_nonlinear_2000}) that have been developed in the context of statistics and machine learning applications, in  truly high-dimensional problems in computational physics \cite{ganapathysubramanian_non-linear_2008,xing_reduced_2015,xing_manifold_2016}.
   One set of applications which really pushes the limits of existing uncertainty propagation techniques, as well as being of significant engineering interest, involves random heterogeneous media \cite{Torquato2001}.    
The macroscale properties of composites (e.g. fiber-reinforced) or polycrystalline materials (e.g. alloys) depend strongly on  the underlying microstructure. The latter is characterized by significant randomness which invariably implies gigantic numbers of random variables \cite{koutsourelakis_probabilistic_2006} and must be propagated across different length scales \cite{panchal_key_2013} in the context of simulation-based analysis and design \cite{olson_designing_2000, yip_handbook_2005}.
Despite recent significant  progress  in the development  of hierarchical \cite{miehe_homogenization_2002} and concurrent \cite{mcdowell_concurrent_2008} { deterministic} multiscale methodologies, most formulations rely on scale separation arguments and the existence of Representative Volume Elements (RVE). However their size, the boundary conditions that must be employed on the RVE in order to extract effective properties are not necessarily uniquely determined nor is their effect in the macroscale response \cite{ostoja-starzewski_microstructural_2010}. Furthermore, only a small portion of this work has been directed to {\em stochastic/probabilistic} multiscale problems \cite{chernatynskiy_uncertainty_2013} and even less, to strategies that would be applicable to high-dimensional, non-Gaussian uncertainties encountered in materials problems \cite{sundararaghavan_multi-length_2008,matous_review_2017}.
   
In this paper we propose a  Bayesian formulation for the construction of reduced-order descriptions for PDE-based models, capable of dealing with high-dimensional stochastic inputs in the coefficients as is the case for example in random media or problems which are characterized by stochastic spatial variability.
It consists of two basic ingredients: a) a (latent) coarse-grained version of the full-order PDE, and b)  a (latent) coarse-to-fine map that relates the outputs of the two models.
We note that coarse-grained   models serve as a stencil for the construction of the reduced description  that  retain  a priori  the salient physical features of the full-order  description.  They are parametrized by a lower-dimensional set of variables which   provide localized,  predictive summaries of the underlying high-dimensional random input. Such a model unavoidably compromises the informational content of the stochastic full-order model and is in general  incapable of providing perfect predictions.   To that end, it is complemented by a probabilistic map that relates  the outputs of the coarse-grained model to the desired outputs of the full-order one. In contrast to existing techniques that perform the dimensionality reduction of the input and the construction of the emulator to the output in two separate steps \cite{ma_kernel_2011}, both of these components are trained {\em simultaneously} in the framework advocated. As a result it is ensured that only low-dimensional features of the input that are predictive of the response (and not of the input itself) are learned and retained.

We employ a Stochastic Variational Inference scheme \cite{paisley_variational_2012,hoffman_stochastic_2013} in order to train the proposed model. This is combined with appropriate prior specifications that promote the discovery of a sparse set of features that maximally compress the random input \cite{Faul2001}. The hierarchical nature of the model allows it to learn from a limited number of full-order runs (in the examples performed these range from 10 to 100).  Its Bayesian nature yields probabilistic predictions of the full-order outputs (independently of  their dimension) that reflect not only the unavoidable information loss mentioned earlier, but also the effect of learning from a finite (and small) dataset.

The remainder of the paper is organized as follows: In Section \ref{sec:methodology}, we present the essential ingredients and provide algorithmic details for the inference and learning processes. In Section \ref{sec:examples}, we present numerical illustrations in the context of high-dimensional elliptic, stochastic PDEs and conclude in Section \ref{sec:conclusions} with some possible extensions involving adaptive refinement and the use of multiphysics models.

\section{Methodology}
\label{sec:methodology}

In general, we use the subscript `$f$' to denote quantities pertaining to the (high-dimensional) full-order model and the subscript `$c$' for quantities associated with the (lower-dimensional) coarsened/reduced-order description. 
We begin with the presentation of the full-order model (FOM) and subsequently explain the essential ingredients of the proposed formulation.

\subsection{SPDE's with random coefficients and the full-order model}
\label{sec:randomCoefficients}

In the modeling of physical systems, material properties such as electrical or thermal conductivity, elastic moduli or fluid permeability are only known up to a stochastic level. 
\revone{We denote 
 by $\lambda(\bs x)$ a scalar (without loss of generality), random field describing any of these properties where $\bs{x}$ is the spatial variable in the problem domain $D$ and consider a governing PDE of the form}
\be
    \mathcal A(\bs x, \lambda(\bs x)) u(\bs x, \lambda(\bs x)) =0, \qquad \tm{for~~} \bs x \in D
\label{spde}
\ee
\revone{where $ \mathcal A(\bs x, \lambda(\bs x))$ is some differential operator (to be specialized in Section \ref{sec:examples}) and $u(\bs x;  \lambda(\bs x))$ is the sought solution field.
Since the method proposed is data-driven, we will not be concerned with the particulars of the solution of the governing equations which are generally complemented with appropriate boundary conditions. We simply make use of the discretized versions  $\bsl_f \in \RR^{N_{el, f}}$ and $\bs{u}_f \in \RR^{N_{dof, f}}$ of the coefficient random field $\lambda(\bs x)$ and the solution $u(\bx)$, respectively.}
We also denote by  $\bs{u}_f(\bsl_f)$ the deterministic map implied by the solution of the discretized PDE which gives the solution vector for each $\bsl_f$. We note that the scale of spatial variability of $\lambda(\bx)$ in many random media necessitates sufficiently fine discretizations of the governing PDE in order to accurately represent the solution. As a consequence, the resulting algebraic system of equations is high-dimensional and cumbersome to solve repeatedly. In the cases considered, both the dimensions of the random input and solution vectors $\bs \lambda_f$, $\bs u_f$ are thus assumed high, i.e. $N_{el, f}$, $N_{dof, f}>>1$. 

\subsection{A Bayesian reduced-order model}

Any attempt to construct an emulator of the input-output map $\bs{u}_f(\bsl_f)$ on the basis of a finite set $\mathcal D = \left\{\bs \lambda_f^{(n)}, \bs u_f^{(n)}\right\}_{n = 1}^N$ of FOM  evaluations is faced with the following difficulties:
\bi
\item the high input dimension $N_{el, f} = \dim(\bs \lambda_f)$ corresponding to the fine scale discretization of the coefficient random field $\lambda(\bs x)$ in relation to the available data $N$. This is known as the ``large $p$, small $N$'' paradigm in statistics \cite{west_bayesian_2003} where $p$ refers to $N_{el, f}=\tm{dim}(\bsl_f)$;
\item the prohibitive cost of enlarging the data set size $N$; and
\item the high dimension $N_{dof, f} = \dim (\bs u_f)$ of the discretized \revtwo{solution}/output vector $\bs{u}_f$.
\ei
It is therefore imperative to employ emulators that encode as much as possible a priori information from the FOM  which, as such, do not require data to be learned. Secondly, it is essential to identify a low-dimensional set of features of the input $\bsl_f$ that are nevertheless predictive of the output \cite{wipf_sparse_2004} and can be learned from the few data available. In the context of deterministic materials' microstructures for example, several upscaling tools have been developed which substitute the high-dimensional microstructures by a low-dimensional set of effective properties \cite{Arbogast2006,efendiev_multiscale_2007}.  Thirdly, it is important to enable effective dimensionality reductions of the output $\bu_f$ that are seamlessly incorporated with the previous two aspects. 
%
%
%

\begin{figure}[t]
  \centering
  \includegraphics[width=.9\textwidth]{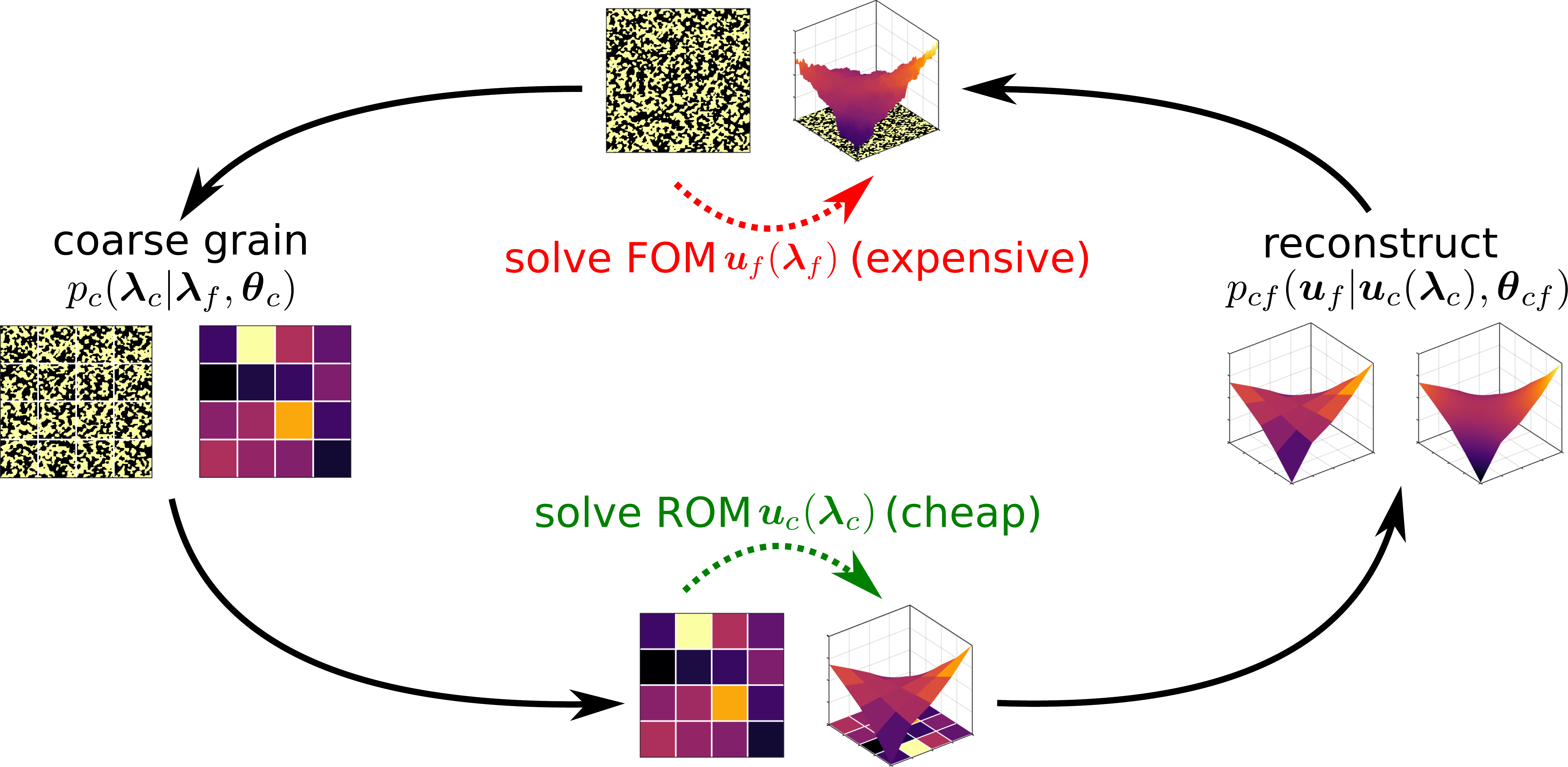}
  \caption{Schematic representation of the model defined by Equation \eqref{bayesnet}. Starting from the top left: In the first step, an effective representation of the FOM input $\bs \lambda_f \mapsto \bs \lambda_c$ is found. Next, the PDE is solved using a (much) coarser discretization. Finally, the FOM solution vector $\bs u_f$ is reconstructed from the coarse one, $\bs u_c \mapsto \bs u_f$.}
  \label{fig:schematic}
\end{figure}

We propose a three-component reduced-order model (ROM) that encapsulates the aforementioned desiderata  and consists of the following steps (Figure \ref{fig:schematic} \cite{Grigo2017}):
\begin{itemize}
    \item a probabilistic mapping from the high-dimensional $\bs \lambda_f$ to a lower-dimensional, coarse-grained representation $\bs \lambda_c$ ($\dim(\bs \lambda_c) \ll \dim(\bs \lambda_f)$). This mapping is mediated by the density $p_c(\bs \lambda_c|\bs \lambda_f, ~\bs \theta_{c})$ parametrized by $\bt_c$;
    
    \item a coarser discretization of the original PDE where $\bs u_c$ is the solution vector ($\tm{dim}(\bu_c)\ll \tm{dim}(\bu_f)$). We denote by $\bu_c(\bsl_c)$ the deterministic input-output mapping implied by this model; and
    \item a probabilistic coarse-to-fine mapping from the output $\bu_c$ of the coarse model to the output of the FOM $\bu_f$. We denote this with the density  $p_{cf}(\bs u_f| \bs u_c, \bs \theta_{cf})$ which is parametrized by $\bt_{cf}$.
\end{itemize}

\begin{figure}[!t]
  \centering
  \includegraphics[width=.5\textwidth]{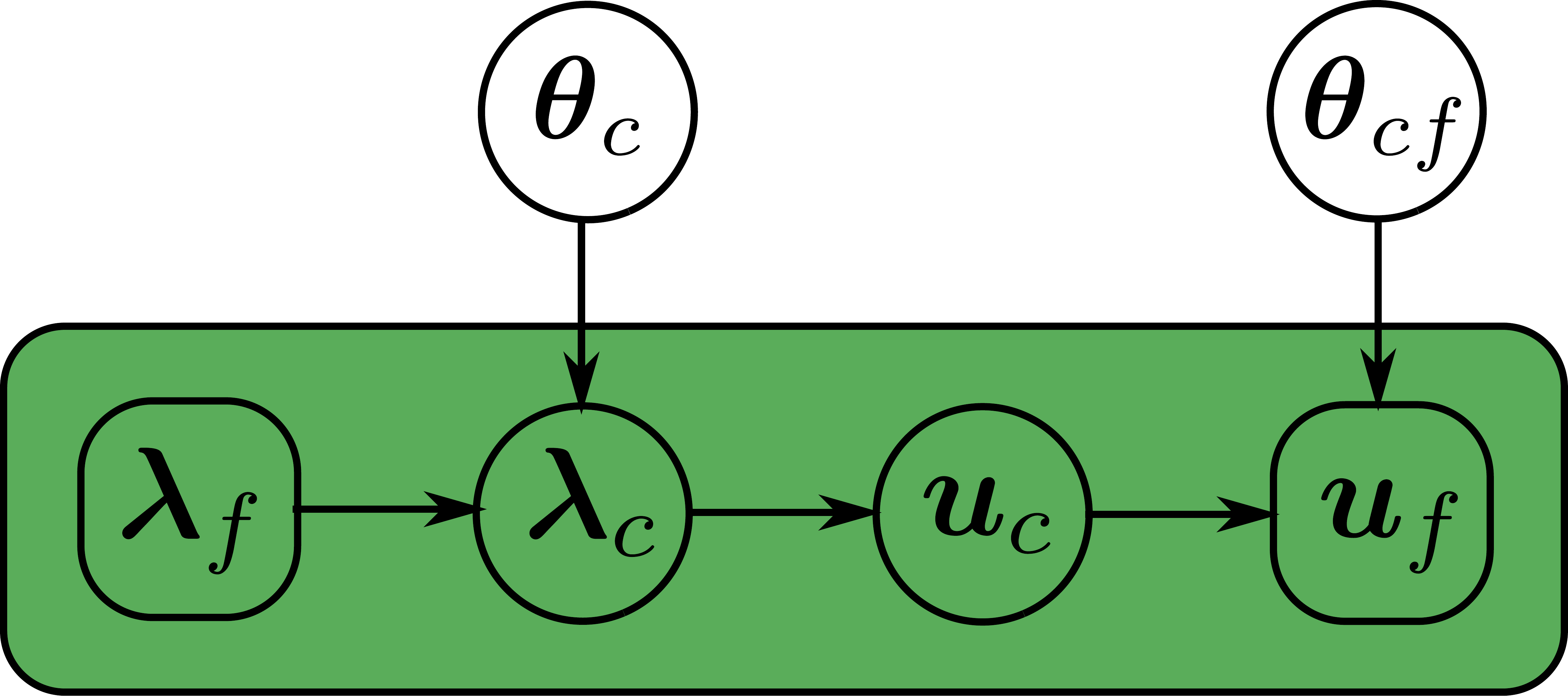}
  \caption{Graphical representation of the three-component Bayesian network implied by $\bar{p}(\bs u_f| \bs \lambda_f, \bs \theta_{cf}, \bs \theta_c)$  in \refeq{bayesnet}. The internal vertices $\bs \lambda_c$, $\bs u_c$ are latent variables. 
  }
  \label{fig:bayesnet}
\end{figure}

The combination of these three components yields the following conditional density:
\begin{equation}
\begin{split}
\bar{p}(\bs u_f| \bs \lambda_f, \bs \theta_{cf}, \bs \theta_c) &= \int \underbrace{p_{cf}(\bs u_f|\bs u_c, \bs \theta_{cf})}_{\tm{decoder}} \underbrace{p_{cm}(\bs u_c| \bs \lambda_c)}_{\tm{coarse model}} \underbrace{p_{c}(\bs \lambda_c| \bs \lambda_f, \bs \theta_c)}_{\tm{encoder}} d\bs u_c d\bs \lambda_c \\
&= \int p_{cf}(\bs u_f|\bs u_c(\bs \lambda_c), \bs \theta_{cf}) p_c(\bs \lambda_c|\bs \lambda_f, \bs \theta_c) d\bs \lambda_c,
\end{split}
\label{bayesnet}
\end{equation}
where we used the fact that  $p_{cm}(\bs u_c| \bs \lambda_c) = \delta(\bs u_c - \bs u_c(\bs \lambda_c))$. The combination of the latent (unobserved) variables $\bsl_c, \bu_c$ with the model parameters $\bt_c,\bt_{cf}$ yield a probabilistic  graphical model \cite{koller_probabilistic_2009} which is formally depicted in Figure \ref{fig:bayesnet}.

The latent variables $\bs \lambda_c$ can be interpreted as a probabilistic filter (encoder) on the FOM  input $\bs \lambda_f$. By solving the coarse model, these are inexpensively transformed to $\bu_c$  which  are finally decoded to predict the FOM output $\bu_f$.  
It is important to note that in order for $\bar{p}(\bs u_f|\bs \lambda_f, \bs \theta_{cf}, \bs \theta_c)$ to approximate well the reference density $p_{\tm{ref}}(\bu_f | \bsl_f)=\delta (\bu_f-\bs u_f(\bs \lambda_f))$, it is irrelevant if the latent variables $\bs \lambda_c$ provide a high-fidelity encoding of $\bs \lambda_f$ in the sense of being able to reconstruct $\bs \lambda_f$. Rather, $\bs \lambda_c$ must be predictive (through $\bu_c$) of  the FOM  response $\bs u_f$.  {\em Hence the $\bs \lambda_c$ implied in our model might be very different from the reduced coordinates identified by a (non)linear dimensionality reduction tool applied directly on $\bsl_f$ (or samples thereof) \cite{Tishby2000}.}

Furthermore, we remark that, in general, and if  no redundancies in $\bs \lambda_f$ are present,  the coarse-graining process effected in the proposed model  will   unavoidably result in  some information loss, i.e. for $\tm{dim}(\bs \lambda_c) << \tm{dim}(\bs \lambda_f)$ there is an upper bound on the mutual information $I(\bs \lambda_c, \bs \lambda_f) \le I_0$. Consequently, there will be uncertainty in the predictions produced by the ROM which we attempt to capture with the aforementioned densities.  We note that this source of uncertainty is independent of the uncertainty arising from the finite dataset which we account for in a Bayesian formulation as discussed in the sequel. 

The decoding density $p_{cf}(\bs u_f| \bs u_c(\bs \lambda_c), \bs \theta_{cf})$ maps the coarse  response vector $\bs u_c$ to its fine-scale counterpart $\bs u_f$, where $\tm{dim}(\bs u_c) \ll \tm{dim}(\bs u_f)$. As  a result, $p_{cf}$  plays the role of a generative model  for dimensionality reduction \cite{tipping_probabilistic_1999} of the FOM output. While many other possibilities exist, given the spatial character of the problems considered, one would expect that this component plays the role of  an interpolant, i.e. it attempts to reconstruct  each  $u_{f, i}$ associated with point $\bs x_i$  by employing the coarse-model outputs $u_{c, j}$, potentially associated with points $\bs x_j$ in the vicinity of $\bs x_i$.

We finally note that the coarse model  $\bs{u}_c(\bsl_c)$ is used as the central building block  of the reduced-order model constructed. Its form determines to a large extent the meaning of the latent variables $\bsl_c$ employed and their association with $\bsl_f$ through $p_c$. Apart from the necessary requirement that it is much less expensive to evaluate than the FOM, one could envisage in its place models accounting for different physics than the FOM, or parametrized models as in the case of reduced-basis techniques (where these parameters would need to be trained in conjunction with $\bt_c,\bt_{cf}$) or even stochastic models (in which case the full $p_{cm}$ would need to be employed in \refeq{bayesnet}).  
%

In the sequel we discuss the specifics of the building components and of the densities $p_c$,  $p_{cf}$ in particular.

%

\subsection{The coarse-graining distribution \texorpdfstring{$p_c$}{}}
\label{sec:p_c}
We denote by $k$ the index of each macro-cell or macro-element in the discretization of the coarse model (see Figure \ref{fig:schematic}). We postulate a relationship of the form\footnote{Often, there are physical bounds of type $\lambda > 0$ or $\lambda_{\tm{lo}} \le \lambda \le \lambda_{\tm{hi}}$ on the random field $\lambda = \lambda(\bs x, \xi(\bs x))$. This should be reflected in the regression model on $\bs \lambda_c$ and can be realized with a link function $\lambda_{c,k} = \chi(z_k)$ where $\chi: \mathbb R \mapsto D_{\lambda}$ with $D_{\lambda}$ the admissible domain for $\lambda$. In such a  case all instances of  $\lambda_{c,k}$ in the subsequent equations should be substituted by $z_k$.} 
\begin{equation}
    \lambda_{c,k} = \sum_{j = 1}^{N_{\tm{features}}} \tilde{\theta}_{c, jk} \varphi_{jk}(\bs \lambda_f) + \sigma_{c,k} Z_k, \qquad Z_k \sim \mathcal N(0, 1),
    \label{linearModelp_c}
\end{equation}
where $\bs{\varphi}_k(\bsl_f)=\{\varphi_{jk}(\bsl_f)\}_{j=1}^{N_{\tm{features}}}$ is a set of predefined feature functions which attempt to filter relevant information of $\bs \lambda_f$ in order to find a $\bs \lambda_{c}$ which is most predictive for the reconstruction of $\bs u_f$. These are combined with  weights $\tilde{\bt}_{c,k}=\{ \tilde{\theta}_{c, jk} \}_{j=1}^{N_{\tm{features}}}$  and a residual noise with variance $\sigma_{c,k}$ which represents the uncertainty in $\lambda_{c,k}$
. The resulting $p_c$ is
\begin{equation}
     p_c(\bs \lambda_c|\bs \lambda_f, \bs \theta_c) = \prod_{k=1}^{\tm{dim}(\bsl_c)} \mathcal{N}( \lambda_{c,k}|~\tilde{\bs \theta}_{c,k}^T \bs{ \varphi}_k(\bs \lambda_f)  , \sigma_{c,k}^2),
    \label{p_c}
\end{equation}
hence $\bt_c=\{ \tilde{\bs \theta}_{c,k}, \sigma_{c,k}^2\}_{k=1}^{\tm{dim}(\bsl_c)}$\footnote{We also denote by $\bs{\Sigma}_c=\tm{diag}(\bs \sigma_{c}^2)$ whenever this is more convenient.}. Naturally, different numbers of feature functions $N_{\tm{features}}$ can be employed for each $k$.  
Using suitable features is a crucial aspect of the expressivity of the model. We provide a detailed list in Appendix \ref{sec:featureList} and note that these consist of various statistical descriptors. Some of these   convey  physical information of the problem, i.e. they should include topological descriptors \cite{Lu1992, Torquato1982, Lowell2006} as well as homogenization-based quantities \cite{Michel1999, Torquato2001}. Others however are based on image recognition tools \cite{Soille1999} or even autoencoder representations \cite{Bengio2009, Tipping1998}. We finally note that employing large numbers of feature functions (as we do in this study) poses important model selection issues which we discuss in Section \ref{sec:prior}.

\subsection{The coarse-to-fine map \texorpdfstring{$p_{cf}$}{}}
\label{sec:p_cf}
This provides a generative interpretation of high-dimensional output $\bu_f$ by employing the (latent) coarse model output $\bu_c$ as  shown schematically in  the third step of Figure \ref{fig:schematic}. In this study, we employ a linear model of the form
\begin{equation}
    p_{cf}(\bs u_f| \bs u_c(\bs \lambda_c), \bs \theta_{cf}) = \mathcal N(\bs u_f| \bs W \bs u_c + \bs b, \bs S),
    \label{pcf}
\end{equation}
where we denote the model parameters $\bs \theta_{cf} =\{ \bs{W}, \bs{b}, \bs{S}\}$.  We note that  $\bs b \in \RR^{\tm{dim}(\bu_f)}$ is a bias vector,  $\bs W \in \RR^{\tm{dim}(\bu_f) \times \tm{dim}(\bu_c)}$ is a projection matrix and $\bs S$ the covariance. To ensure that the number of unknown parameters scales linearly with the dimension of the FOM output $\bu_f$, we employ a diagonal $\bs S$. Furthermore, and in order to reduce the amount of data needed, we exploit the spatial characteristics of the problem in order to restrict the number of free parameters in  $\bs W, \bs b$  as discussed in Section \ref{sec:examples}. 


\subsection{Model training}
\label{sec:modelTraining}
Given the aforementioned components of the proposed model, we discuss the calibrations  of the model parameters $\bs \theta = \{\bs \theta_{cf}, \bs \theta_c\}$ on the basis of a set of $N$  FOM observations $\mathcal D = \left\{\bs \lambda_f^{(n)}, \bs u_f^{(n)} \right\}_{n = 1}^N$. Following the Bayesian paradigm, the plausibility for a certain parameter value $\bs \theta$ is given by the posterior
\begin{equation}
p(\bs \theta|\mathcal D) \propto \mathcal L(\mathcal D|\bs \theta) p(\bs \theta),
\end{equation}
where $p(\bs \theta)$ is a model prior to be specified and
\begin{equation}
\mathcal L(\mathcal D| \bs \theta) = \prod_{n = 1}^N \bar{p}(\bs u_f^{(n)}|\bs \lambda_f^{(n)}, \bs \theta)
\label{likelihood1}
\end{equation}
is the likelihood function. We note that maximizing the log-likelihood with respect to $\bt$ is equivalent to minimizing the Kullback-Leibler divergence \cite{bishop_pattern_2007} between the reference density $p_{\tm{ref}}(\bu_f | \bsl_f)=\delta (\bu_f-\bs u_f(\bs \lambda_f))$ and the model-implied density $\bar{p}(\bs u_f|\bs \lambda_f, \bs \theta_{cf}, \bs \theta_c)$ in \eqref{bayesnet}. The latter however implies an integration w.r.t. $\bs \lambda_c$ which despite the form of $p_c$ and $p_{cf}$ is analytically intractable due to the dependence on the coarse model output $\bs u_c(\bs \lambda_c)$.
Furthermore, due to the dimensionality of the model parameters (particularly $\bt_{cf}$) we adopt a hybrid strategy which is based on the computation of the Maximum a Posteriori estimate $\bt_{\tm{MAP}}$ of $\bs \theta$,
\begin{equation}
\bs \theta_{\tm{MAP}} = \arg\max_{\bs \theta} p(\bs \theta|\mathcal D)
\label{maxpost}
\end{equation}
and the use of Laplace approximations to approach the true posterior \cite{mackay_information_2003}.
Hence, in Section \ref{sec:MAP} we  put forth a Variational Expectation-Maximization scheme \cite{beal_variational_2003} for the efficient computation of $\bt_{\tm{MAP}}$. Particular aspects that pertain to the prior specifications are presented in Section \ref{sec:prior} and in Section \ref{sec:modelpred} the use of the trained  model in producing probabilistic predictive estimates is 
 discussed.

\subsubsection{Maximizing the posterior}
\label{sec:MAP}
Equations (\ref{bayesnet}) and (\ref{likelihood1}) lead to
\begin{equation}
p(\bs \theta| \mathcal D) \propto p(\bt) \cdot \prod_{n = 1}^N \int p_{cf}(\bs u_f^{(n)}|\bs u_c(\bs \lambda_c^{(n)}), \bs \theta_{cf}) p_c(\bs \lambda_c^{(n)}|\bs \lambda_f^{(n)}, \bs \theta_c) d\bs \lambda_c^{(n)},
\label{posterior1}
\end{equation}
where $p(\bt)$ denotes the prior on the model parameters $\bt = \{\bs \theta_{cf}, \bs \theta_c \}$.
In order to carry out the maximization of the intractable objective we resort to the \tit{Expectation-Maximization} (EM) algorithm \cite{Dempster1977}. Based on  Jensen's inequality,
 we can lower-bound the $\log$ likelihood $\mathcal L(\mathcal D| \bs \theta_{cf}, \bs \theta_c)$ \eqref{likelihood1} as
\begin{equation}
    \begin{split}
        \log \mathcal L(\mathcal D| \bs \theta_{cf}, \bs \theta_c) &= \sum_{n = 1}^N \log \int p_{cf}(\bs u_f^{(n)}|\bs u_c(\bs \lambda_c^{(n)}), \bs \theta_{cf}) p_c(\bs \lambda_c^{(n)}|\bs \lambda_f^{(n)}, \bs \theta_c) d\bs \lambda_c^{(n)} \\
        &\ge \sum_{n = 1}^N \int q_n(\bs \lambda_c^{(n)}) \log \left(\frac{p_{cf}(\bs u_f^{(n)}|\bs u_c(\bs \lambda_c^{(n)}), \bs \theta_{cf}) p_c(\bs \lambda_c^{(n)}|\bs \lambda_f^{(n)}, \bs \theta_c)}{q_n(\bs \lambda_c^{(n)})} \right) d\bs \lambda_c^{(n)} \\
        &= \sum_{n = 1}^N \mathcal F^{(n)}(q_n(\bs \lambda_c^{(n)}); \bs \theta) = \mathcal F(\left\{q_n(\bs \lambda_c^{(n)}) \right\}_{n = 1}^N; \bs \theta),
    \end{split}
    \label{eq:vlb}
\end{equation}
where $q_n(\bs \lambda_c^{(n)})$ are arbitrary probability densities.
Consequently, the $\log$ posterior \eqref{posterior1} has the lower bound
\begin{equation}
    \log p(\bs \theta|\mathcal D) \ge \mathcal F(\left\{q_n(\bs \lambda_c^{(n)}) \right\}_{n = 1}^N; \bs \theta) + \log p(\bs \theta).
    \label{posteriorLB}
\end{equation}

The basic idea behind the EM-algorithm is to maximize iteratively the lower-bound \eqref{posteriorLB}  with respect to parameters $\bs \theta$ and the auxiliary distributions $\left\{q_n(\bs \lambda_c^{(n)}) \right\}_{n = 1}^N$. One can readily verify that for a  given value of  $\bs \theta = \bs \theta^{(t)}$, the optimal $q_n$'s are given by the posterior of each $ \lambda_c^{(n)}$, i.e.
\begin{equation}
    q^{\tm{opt}}_n(\bs \lambda_c^{(n)}) = p_n(\bs \lambda_c^{(n)}| \bs \theta^{(t)}) \propto p_{cf}(\bs u_f^{(n)}|\bs u_c(\bs \lambda_c^{(n)}), \bs \theta_{cf}^{(t)}) p_c(\bs \lambda_c^{(n)}|\bs \lambda_f^{(n)}, \bs \theta_c^{(t)}).
    \label{Estep1}
\end{equation}
In this case the lower-bound becomes tight and the inequality in \eqref{posteriorLB} turns into an equality. The previous suggests the following maximization process whereby at each iteration $t$ one alternates between:
\vspace{2mm}
\begin{description}
    \item[\tbf{E-step:}] Given the current parameter values $\bs \theta^{(t)}$, find the $q^{(t + 1)}_n(\bs \lambda_c^{(n)})$ that maximize \\ $\mathcal{F}(\left\{q_n(\bs \lambda_c^{(n)}) \right\}_{n = 1}^N; \bs \theta^{(t)})$ (see \refeq{Estep1}).  
    \vspace{2mm}
    \item[\tbf{M-step:}] Given the current expected values $\left< ~.~\right>_{q_n^{(t + 1)}}$, maximize the posterior lower bound
    \begin{equation}
        \bt^{(t + 1)}= \arg\max_{\bs \theta} \left(\mathcal F(\left\{q_n^{(t + 1)}(\bs \lambda_c^{(n)}) \right\}_{n = 1}^N; \bs \theta) + \log p(\bs \theta) \right)
    \end{equation}
    to find the next best estimates $\bs \theta^{(t + 1)}$. 
\end{description}
\vspace{2mm}
The  iterations are repeated until a suitable convergence criterion on the parameters $\bt$ is met.
Partial or incomplete updates can readily be performed and could potentially lead to computational benefits \cite{neal_view_1998}.

\paragraph{Stochastic Variational Inference during the E-step}
We emphasize  that no further FOM runs (apart from those performed to generate the training data $\mathcal{D}$) are needed in any of the steps above but note that the E-step is analytically intractable due to the dependence on the  coarse model outputs $\bs u_c(\bs \lambda_c^{(n)})$.  In order to avoid employing Monte Carlo sampling schemes (e.g. MCMC, SMC) which, despite the unbiased estimates they produce, are not as efficient in terms of the number of times $\bs u_c(\bs \lambda_c)$ needs to be evaluated, we propose employing an approximate inference scheme that relies on Stochastic Variational Inference (SVI)  \cite{paisley_variational_2012,hoffman_stochastic_2013}. These yield sub-optimal approximations to the densities needed in the E-step which are nevertheless shown to be sufficient for  accurate estimation of $\bt_{\tm{MAP}}$ \cite{Blei2017}.  
To that end, we employ a family of densities $q_{n, \bs \xi_n}(\bs \lambda_c^{(n)})$ parametrized by  $\bs \xi_n$ and seek their optimal values in terms of maximizing the variational lower-bound $\mathcal{F}^{(n)}$. In particular, at each iteration (i.e. given $\bt^{(t)}$) and for each $n$, we seek\footnote{It can be shown that the optimization problem in \refeq{eq:svi} is equivalent to minimizing the Kullback-Leibler divergence between $q_{n, \bs \xi_n}(\bs \lambda_c^{(n)})$ and $q^{opt}_n(\bs \lambda_c^{(n)})$ given in \refeq{Estep1}.}
\be
\bs{\xi}_n = \arg \max_{\bs{\xi}_n}  \mathcal{F}^{(n)}_{VI}(\bs{\xi}_n)
\ee
where
\begin{equation}
\begin{split}
 \mathcal{F}^{(n)}_{VI}(\bs{\xi}_n)  & =  \mathcal F^{(n)}(q_{n, \bs \xi_n}(\bs \lambda_c^{(n)}); \bs \theta) \\
&=\int q_{n, \bs \xi_n}(\bs \lambda_c^{(n)}) \log \left(\frac{p_{cf}(\bs u_f^{(n)}|\bs u_c(\bs \lambda_c^{(n)}), \bs \theta_{cf}) p_c(\bs \lambda_c^{(n)}|\bs \lambda_f^{(n)}, \bs \theta_c)}{q_{n, \bs \xi_n}(\bs \lambda_c^{(n)})} \right) d\bs \lambda_c^{(n)} \\
  & =  \left<\log p_{cf}(\bs u_f^{(n)}|\bs u_c(\bs \lambda_c^{(n)})\right>_{ q_{n, \bs \xi_n}(\bs \lambda_c^{(n)})} + \left<\log p_c(\bs \lambda_c^{(n)}|\bs \lambda_f^{(n)}, \bs \theta_c)\right>_{q_{n, \bs \xi_n}(\bs \lambda_c^{(n)})} \\
 & +H(q_{n, \bs \xi_n}(\bs \lambda_c^{(n)}))
\label{eq:svi}
\end{split}
\end{equation}
where $\left<~ . ~\right>_{q_{n, \bs \xi_n}(\bs \lambda_c^{(n)})}$ imply expectations with respect to $q_{n, \bs \xi_n}(\bs \lambda_c^{(n)})$ and $H(q_{n, \bs \xi_n}(\bs \lambda_c^{(n)})$ is the corresponding Shannon entropy.
Since the derivatives of the objective above with respect to $\bs{\xi}_n$ involve expectations with respect to $q_{n, \bs \xi_n}(\bs \lambda_c^{(n)})$ and in order to minimize the variance in these estimates, we employ the reparametrization trick \cite{Kingma2013}. In particular, for the family of multivariate Gaussians $q_{n, \bs \xi_n}(\bs \lambda_c^{(n)})$ $ = \mathcal N(\bs \lambda_c^{(n)}| \bs \mu_{VI}^{(n)}, \bs \Sigma_{VI}^{(n)})$ where $\bs \xi_n = \{ \bs \mu_{VI}^{(n)},  \bs \Sigma_{VI}^{(n)} \}$\footnote{We use diagonal covariances $\bs \Sigma_{VI}^{(n)}$.}, the reparametrization trick consists of expressing  $\bsl^{(n)}=  \bs \mu_{VI}^{(n)}+\sqrt{ \bs \Sigma_{VI}^{(n)}} \bs{\epsilon}^{(n)}$ where $ \bs{\epsilon}^{(n)} \sim \mathcal{N}(\bs{0},\bs{I})$. Upon substitution in the objective of \eqref{eq:svi}, we obtain:
\begin{equation}
\begin{split}
\mathcal{F}^{(n)}_{VI}(\bs{\xi}_n) & = \left<\log p_{cf}(\bs u_f^{(n)}|\bs u_c( \bs \mu_{VI}^{(n)}+\sqrt{\bs \Sigma_{VI}^{(n)} } \bs{\epsilon}^{(n)}))\right>_{ \mathcal{N}(\bs{\epsilon}^{(n)}|\bs{0},\bs{I} )}  \\
 & + \left<\log p_c( \bs \mu_{VI}^{(n)} + \sqrt{ \bs \Sigma_{VI}^{(n)} } \bs{\epsilon}^{(n)} |\bs \lambda_f^{(n)}, \bs \theta_c)\right>_{\mathcal{N}(\bs{\epsilon}^{(n)}|\bs{0},\bs{I} )} \\
& +H(q_{n, \bs \xi_n}(\bs \lambda_c^{(n)})).
\label{eq:svi2}
\end{split}
\end{equation}
Given that (up to a constant) $H(q_{n, \bs \xi_n}(\bs \lambda_c^{(n)}))=\frac{1}{2}\log |\bs \Sigma_{VI}^{(n)}|$ and after application of the chain rule we obtain the following derivatives:
\be
\begin{split}
\cfrac{ \pa \mathcal{F}^{(n)}_{VI}}{\pa \bs \mu_{VI}^{(n)}} &= \left< \cfrac{\pa \log p_{cf}}{\pa \bs{u}_c} \cfrac{\pa \bu_c}{\pa \bsl_c} \right>_{ \mathcal{N}(\bs{\epsilon}^{(n)}|\bs{0},\bs{I} )}  +
 \left< \cfrac{\pa \log p_c}{\pa \bsl_c}\right>_{ \mathcal{N}(\bs{\epsilon}^{(n)}|\bs{0},\bs{I} )}  \\
 \cfrac{ \pa \mathcal{F}^{(n)}_{VI}}{\pa \sqrt{\bs \Sigma_{VI}^{(n)} }} &  = \left< \cfrac{\pa \log p_{cf}}{\pa \bs{u}_c} \cfrac{\pa \bu_c}{\pa \bsl_c} (\bs{\epsilon}^{(n)})^T\right>_{ \mathcal{N}(\bs{\epsilon}^{(n)}|\bs{0},\bs{I} )}  +
 \left< \cfrac{\pa \log p_c}{\pa \bsl_c} (\bs{\epsilon}^{(n)})^T\right>_{ \mathcal{N}(\bs{\epsilon}^{(n)}|\bs{0},\bs{I} )}  +(\bs \Sigma_{VI}^{(n)})^{-1}.
\end{split}
\ee
If not given in closed form, the expectations above with respect to $\bs{\epsilon}^{(n)}$ are estimated with Monte Carlo and the (noisy) derivatives are used to update $\bs{\xi}_n$ in conjunction with the \tit{ADAM} stochastic optimization method \cite{Kingma2014}. We note finally that the gradients above involve   derivatives of the coarse model's output w.r.t. the coefficients $\bs \lambda_c$, $\frac{\pa \bs u_c}{\pa \bs \lambda_c}$. These can efficiently be obtained given the size of the model by solving the adjoint equations (see e.g. \cite{Heinkenschloss2008}).

\paragraph{M-step: model parameter updates}
For maximization of the posterior lower bound, we use gradients of $\mathcal F$ from \refeq{eq:vlb},
\begin{align}
    \nabla_{\bs \theta_{cf}} \mathcal F(\left\{q_n^{(t + 1)}(\bs \lambda_c^{(n)}) \right\}_{n = 1}^N; \bs \theta_{cf}, \bs \theta_c) &= \sum_{n = 1}^N \left< \nabla_{\bs \theta_{cf}} \log p_{cf}(\bs u_f^{(n)}|\bs u_c(\bs \lambda_c^{(n)}), \bs \theta_{cf}) \right>_{q_n^{(t + 1)}}, \\
    \nabla_{\bs \theta_{c}} \mathcal F(\left\{q_n^{(t + 1)}(\bs \lambda_c^{(n)}) \right\}_{n = 1}^N; \bs \theta_{cf}, \bs \theta_c) &= \sum_{n = 1}^N \left< \nabla_{\bs \theta_{c}} \log p_c(\bs \lambda_c^{(n)}|\bs \lambda_f^{(n)}, \bs \theta_c) \right>_{q_n^{(t + 1)}}.
\end{align}
Given the model densities $p_{cf}(\bs u_f| \bs u_c(\bs \lambda_c), \bs \theta_{cf})$ (Equation \eqref{pcf}), $p_c(\bs \lambda_c|\bs \lambda_f, \bs \theta_c)$ (Equation \eqref{p_c}), we obtain:
\begin{small}
\begin{align}
    \label{dF_dW}
    \nabla_{\bs W} \mathcal F &= \bs S^{-1}\sum_{n = 1}^N\left((\bs u_f^{(n)} - \bs b) \left<\bs u_c^T(\bs \lambda_c^{(n)}) \right>_{q_n^{(t + 1)}} - \bs W \left<\bs u_c(\bs \lambda_c^{(n)}) \bs u_c^T(\bs \lambda_c^{(n)}) \right>_{q_n^{(t + 1)}} \right), \\
    \nabla_{\bs b} \mathcal F &= \bs S^{-1} \left(\sum_{n = 1}^{N}\left(\bs u_f^{(n)}  - \bs W\left<\bs u_c(\bs \lambda_c^{(n)}) \right>_{q_n^{(t + 1)}}\right) -N\bs b \right), \\
    \label{grad_S}
    \nabla_{\bs S} \mathcal F &= \frac{\bs S^{-1}}{2}\left(\sum_{n = 1}^N \left< (\bs u_f^{(n)} - \bs b - \bs W \bs u_c^{(n)}) (\bs u_f^{(n)} - \bs b - \bs W \bs u_c^{(n)})^T \right>_{q_n^{(t + 1)}} \bs S^{-1} - N\right), \\
        \label{grad_thetac}
    \nabla_{\tilde{\bs \theta}_c} \mathcal F &= \sum_{n = 1}^N \left(\bs \Phi^T(\bs \lambda_f^{(n)}) \bs \Sigma_c^{-1} \left<\bs \lambda_c^{(n)} \right>_{q_n^{(t + 1)}} - \bs \Phi^T(\bs \lambda_f^{(n)}) \bs \Sigma_c^{-1}\bs \Phi(\bs \lambda_f^{(n)})\tilde{\bs \theta}_c \right), \\
    \nabla_{\bs \Sigma_c} \mathcal F &= \frac{\bs \Sigma_c^{-1}}{2} \left(\sum_{n = 1}^N \left<(\bs \lambda_c^{(n)} - \bs \Phi(\bs \lambda_f^{(n)}) \tilde{\bs \theta}_c) (\bs \lambda_c^{(n)} - \bs \Phi(\bs \lambda_f^{(n)}) \tilde{\bs \theta}_c)^T \right>_{q_n^{(t + 1)}}\bs \Sigma_c^{-1} - N \right).
    \label{dF_dSigma}
\end{align}
\end{small}
In a maximum likelihood setting, i.e. with uniform priors $p(\bt) \propto  \tm{const.}$, we observe that the update equations given by $\nabla_{\bs \theta} \mathcal F = 0$ are linear in all parameters $\bs \theta = \{ \bs W ,  \bs b, \bs S,  \tilde{\bs \theta}_c,  \bs \Sigma_c )$ and closed-form updates can be carried out. We provide these update equations in Section \ref{sec:examples} where priors are specified. 
In general, the  gradients above can also be used together with the $\log$ prior gradients in any iterative (stochastic) optimization scheme. \revone{A complexity analysis of training and prediction stages is given in Section \ref{sec:complexity}.}

\subsubsection{Prior specification}
\label{sec:prior}
\begin{algorithm}[t]
\caption{Posterior maximization}
\label{posteriorOpt}
\begin{algorithmic}[1]
\REQUIRE $\bs W^{(0)}$, $\bs b^{(0)}$, $\bs S^{(0)}$, $\tilde{\bs \theta}_c^{(0)}$, $\bs \Sigma_c^{(0)}$, $\bs \gamma^{(0)}$ \hfill\COMMENT{Initialization}
\STATE Set $t\leftarrow 0$
\WHILE{(not converged)}
\STATE \tit{E-step:} \hfill\COMMENT{Completely parallelizable in $n$}
\FOR{$n = 1$ \TO $N$}
\STATE Update $q_n^{(t + 1)}(\bs \lambda_c^{(n)})$ according to \eqref{Estep1}
\STATE Estimate $\left<\bs \lambda_c^{(n)} \right>_{q_n^{(t + 1)}}$, $\left<\bs \lambda_c^{(n)} \bs (\bsl_c^{(n)})^T \right>_{q_n^{(t + 1)}}$, and  $\left<\bs u_c(\bs \lambda_c^{(n)}) \right>_{q_n^{(t + 1)}}$, $\left<\bs u_c(\bs \lambda_c^{(n)}) \bs u_c^T(\bs \lambda_c^{(n)}) \right>_{q_n^{(t + 1)}}$
\ENDFOR
\STATE \tit{M-step:}
\STATE Find $\bs W^{(t + 1)}$, $\bs b^{(t + 1)}$, $\bs S^{(t + 1)}$, $\tilde{\bs \theta}_c^{(t + 1)}$, $\bs \Sigma_c^{(t + 1)}$ by maximization of $\mathcal F$ using \eqref{dF_dW}--\eqref{dF_dSigma}
\STATE \tit{Inner E-step:}
\STATE Given the posterior $q_{\tilde{\bs \theta}_c}^{(t + 1)}(\tilde{\bs \theta}_c)$ in \eqref{posteriorTheta}, estimate $\left<\tilde{\theta}_{c,i}^2 \right>_{q_{\tilde{\bs \theta}_c}^{(t + 1)}}$ using Laplace approximation
\STATE \tit{Inner M-step:}
\STATE Maximize the evidence lower bound $\mathcal G(q_{\tilde{\bs \theta}_c}^{(t + 1)}(\tilde{\bs \theta}_c); \bs \gamma)$ given in \eqref{elboGamma} using update equation \eqref{updateGamma}
\STATE $t \leftarrow t + 1$
\ENDWHILE
\RETURN $\bs W_{\tm{MAP}}$, $\bs b_{\tm{MAP}}$, $\bs S_{\tm{MAP}}$, $\tilde{\bs \theta}_{c, \tm{MAP}}$, $\bs \Sigma_{c, \tm{MAP}}$, $\bs \gamma^*$
\end{algorithmic}
\end{algorithm}
A key point of the proposed model is the discovery of predictive features of the high-dimensional input $\bs \lambda_f$ during the coarse-graining process $\bs \lambda_f \mapsto \bs \lambda_c$. This dimensionality reduction process takes place in the linear model for $p_c$ (Equation \eqref{linearModelp_c}) and  depends on the vocabulary of feature functions $\varphi(\bs \lambda_f)$ employed. One strategy is to sequentially add features from a parametric \cite{Bilionis2012, Pietra1997} or predefined \cite{efron2004, Kohavi1997, Narendra1977} set of feature functions $\varphi(\bs \lambda_f)$ upon optimization of a suitable predictive performance measure. Another way to proceed is to start with a large dictionary of features $\varphi$, which can potentially produce an excessively complex model that overfits and is hampered by non-unique optima.

In order to regularize the problem, we employ a prior on the feature function coefficients $\tilde{\bs \theta}_c$ that favors sparse solutions where only a few components assume non-zero values. Apart from computational advantages (pruned out features do not need to be evaluated for predictions), such a prior reveals the features that are most predictive for $\bs \lambda_c$ and thus may provide further insight to the underlying physics of the problem. Several sparsity enforcing approaches were tested in this work, including the  Laplacian prior (or LASSO regression \cite{Tibshirani1996, Hans2009}) as well as Student-$t$ type prior models \cite{MacKay1996, Neal1996, Figueiredo2003}. We achieved best experimental results using a slightly modified version of the \tit{Relevance Vector Machine} (RVM) \cite{Tipping2000, Tipping2001, Bishop2000} adjusted for  use in latent variable models.

The basic prior model is  of the form $p(\tilde{\bs \theta}_c|\bs \gamma) = \mathcal N(\tilde{\bs \theta}_c| \bs 0, \tm{diag}[\bs \gamma])$ where $\bs \gamma \in \mathbb{R}_+^{N_{\tm{features}}}$ is a vector of non-negative hyperparameters describing the prior variance of each feature component. These hyperparameters are estimated by first integrating out the model parameters $\tilde{\bs \theta}_c$ and then performing what is known as type-II maximum likelihood or evidence maximization \cite{Neal1996}. Given the likelihood \eqref{likelihood1}
\begin{equation*}
    \mathcal L(\tilde{\bs \theta}_c) = \prod_{n = 1}^N \int p_{cf}(\bs u_f^{(n)}| \bs u_c(\bs \lambda_c^{(n)}), \bs \theta_{cf}) p_c(\bs \lambda_c^{(n)}| \bs \lambda_f^{(n)}, \tilde{\bs \theta_c}, \bs \Sigma_c) d\bs \lambda_c^{(n)},
\end{equation*}
the marginal w.r.t. $\tilde{\bs \theta}_c$ is
\begin{equation}
\mathcal P(\bs \gamma) = \int \mathcal L(\tilde{\bs \theta}_c) p(\tilde{\bs \theta}_c|\bs \gamma) d\tilde{\bs \theta}_c.
\end{equation}
We determine the value of the hyperparameters as
\begin{equation}
\bs \gamma^* = \arg\max_{\bs \gamma} \mathcal P(\bs \gamma),
\end{equation}
which is computed in an inner loop  of Expectation-Maximization (EM). To that end, we use the $\log$ evidence lower bound
\begin{align}
\log \mathcal P(\bs \gamma) &= \log \int \mathcal L(\tilde{\bs \theta}_c) p(\tilde{\bs \theta}_c|\bs \gamma) d\tilde{\bs \theta_c} \\
&\ge \int q_{\tilde{\bs \theta}_c}(\tilde{\bs \theta}_c) \log \left(\frac{\mathcal L(\tilde{\bs \theta}_c) p(\tilde{\bs \theta}_c|\bs \gamma)}{q_{\tilde{\bs \theta}_c}(\tilde{\bs \theta}_c)} \right) d\tilde{\bs \theta}_c  = \mathcal G(q_{\tilde{\bs \theta}_c}(\tilde{\bs \theta}_c); \bs \gamma),
\label{elboHyperparam}
\end{align}
where $q_{\tilde{\bs \theta}_c}(\tilde{\bs \theta}_c)$ is an arbitrary auxiliary distribution. The $q_{\tilde{\bs \theta}_c}^{(t + 1)}(\tilde{\bs \theta}_c)$ that maximizes \eqref{elboHyperparam} for a given $\bs \gamma^{(t)}$ (as the inequality becomes an equality) is
\begin{equation}
q_{\tilde{\bs \theta}_c}^{(t + 1)}(\tilde{\bs \theta}_c) \propto \mathcal L(\tilde{\bs \theta}_c) p(\tilde{\bs \theta}_c| \bs \gamma^{(t)}).
\label{posteriorTheta}
\end{equation}
Using the fact that
\begin{equation}
\log p(\tilde{\bs \theta}_c|\bs \gamma) \propto -\frac{1}{2} \sum_{i = 1}^{N_{\tm{features}}} \log \gamma_i - \frac{1}{2}\sum_{i = 1}^{N_{\tm{features}}} \frac{ \tilde{\theta}_{c,i}^2 }{\gamma_i},
\end{equation}
and  keeping only terms that depend on $\bs \gamma$, we get
\begin{equation}
\mathcal G(q_{\tilde{\bs \theta}_c}^{(t + 1)}(\tilde{\bs \theta}_c); \bs \gamma) \propto -\frac{1}{2}\sum_{i = 1}^{N_{\tm{features}}} \log \gamma_i - \frac{1}{2}\sum_{i = 1}^{N_{\tm{features}}} \gamma_i^{-1} \left<\tilde{\theta}_{c,i}^2 \right>_{q_{\tilde{\bs \theta}_c}^{(t + 1)}}.
\label{elboGamma}
\end{equation}
Setting the derivatives $\frac{\pa}{\pa \gamma_j} \mathcal G(q_{\tilde{\bs \theta}_c}^{(t + 1)}(\tilde{\bs \theta}_c); \bs \gamma)$ to $0$ yields the update equations
\begin{equation}
\gamma_j^{(t + 1)} = \left< \tilde{\theta}_{c,j}^2 \right>_{q_{\tilde{\bs \theta}_c}^{(t + 1)}}.
\label{MstepGamma}
\end{equation}
To estimate the expected value $ \left< \tilde{\theta}_{c,j}^2 \right>_{q_{\tilde{\bs \theta}_c}^{(t + 1)}}$, we perform Laplace approximation  
\begin{equation}
q_{\tilde{\bs \theta}_c}^{(t + 1)}(\tilde{\bs \theta}_c) \approx \mathcal N(\tilde{\bs \theta}_c| \tilde{\bs\theta}_c^{(t)}, \tilde{\bs \Sigma}^{(t)}),
\end{equation}
where
\begin{equation}
\tilde{\bs \theta}_c^{(t)} = \arg\max_{\bs \theta_c} \mathcal L(\tilde{\bs \theta}_c) p(\tilde{\bs \theta}_c| \bs \gamma^{(t)})
\end{equation}
is the maximum of $q_{\tilde{\bs \theta}_c}^{(t + 1)}(\tilde{\bs \theta}_c)$  in \eqref{posteriorTheta} for a given $\bs \gamma^{(t)}$, which we find using EM as described in \ref{sec:MAP}. According to Laplace approximation, the covariance $\tilde{\bs \Sigma}^{(t)}$ is given by
\begin{equation}
(\tilde{\bs \Sigma}^{(t)})^{-1} = - \nabla_{\tilde{\bs \theta}_c} \nabla_{\tilde{\bs\theta}_c} \left. \log q_{\tilde{\bs \theta}_c}^{(t + 1)}(\tilde{\bs \theta}_c) \right|_{\tilde{\bs \theta}_c = \tilde{\bs \theta}_c^{(t)}} = \sum_{n = 1}^N \bs \Phi^T(\bs \lambda_f^{(n)}) \bs \Sigma_c^{-1} \bs \Phi(\bs \lambda_f^{(n)}) + (\tm{diag}[\bs \gamma^{(t)}])^{-1},
\end{equation}
where $\nabla_{\tilde{\bs \theta}_c} \nabla_{\tilde{\bs\theta}_c} \left. \log q_{\tilde{\bs \theta}_c}^{(t + 1)}(\tilde{\bs \theta}_c) \right|_{\tilde{\bs \theta}_c = \tilde{\bs \theta}_c^{(t)}}$ denotes the Hessian of $\log q_{\tilde{\bs \theta}_c}^{(t + 1)}(\tilde{\bs \theta}_c)$ at $\tilde{\bs \theta}_c = \tilde{\bs \theta}_c^{(t)}$. We finally get
\begin{equation}
\gamma_j^{(t + 1)} = \left< \tilde{\theta}_{c,j}^2 \right>_{q_{\tilde{\bs \theta}_c}^{(t + 1)}} \approx (\tilde{\theta}_{c, j}^{(t)})^2 + \tilde{\Sigma}_{jj}^{(t)},
\label{updateGamma}
\end{equation}
where the `$\approx$' accounts for the Laplace approximation. After convergence of $\bs \gamma^{(t)}, \tilde{\bs \theta}_c^{(t)}, \tilde{\bs \Sigma}^{(t)}$ to $\bs \gamma^*, \tilde{\bs \theta}_{c, \tm{MAP}}, \tilde{\bs \Sigma}_{\tm{MAP}}$, the posterior on $\tilde{\bs \theta}_c$ is approximated by
\begin{equation}
p(\tilde{\bs \theta}_c|\mathcal D) \approx \mathcal N(\tilde{\bs \theta}_c| \tilde{\bs \theta}_{c, \tm{MAP}}, \tilde{\bs \Sigma}_{\tm{MAP}}).
\label{LaplaceApprox}
\end{equation}
It can be shown \cite{Faul2001, Tipping2001} that many of the prior variance parameters $\gamma_i$ converge to $0$ such that the corresponding features $\varphi_i$ are effectively deactivated.
A summary of the optimization scheme is given in Algorithm \ref{posteriorOpt}.


\subsection{Model predictions}
\label{sec:modelpred}

A key feature of the proposed model is the ability to produce probabilistic predictions that reflect the various sources of uncertainty enumerated previously.  Given the posterior $p(\bt | \mathcal{D})$  on the model parameters $\bt$ which in the case of MAP estimates can be substituted by $\delta(\bt-\bt_{\tm{MAP}})$ and for a new input $\bsl_f$, the Bayesian reduced-order model formulated yields a predictive posterior density $p_{\tm{pred}}(\bu_f | \bsl_f, \mathcal{D})$ for the FOM output $\bu_f$ of the form
\be
\begin{split}
p_{\tm{pred}}(\bu_f | \bsl_f, \mathcal{D})  & = \int p( \bu_f, \bt | \bsl_f, \mathcal{D}) ~d\bt \\
 & = \int  \underbrace{ p( \bu_f | \bsl_f,  \bt)}_{\textrm{ Equation \eqref{bayesnet}} }~ p(\bt| \mathcal{D})~d\bt \\
 &= \int \left( \int p_{cf}(\bs u_f|\bs u_c(\bs \lambda_c), \bs \theta_{cf}) p_c(\bs \lambda_c|\bs \lambda_f, \bs \theta_c) d\bs \lambda_c \right)~ p(\bt| \mathcal{D})~d\bt.
\end{split}
\ee
While the aforementioned density is analytically intractable, samples can inexpensively be generated by following the steps, see also Figure \ref{offline_online_chart}:
%
\begin{itemize}
    \item drawing a sample $\bt = \{\bt_c,\bt_{cf}\}$ from the posterior $p(\bt | \mathcal{D})$; 
    \item drawing a sample $\bs \lambda_c \sim p_c(\bs \lambda_c|\bs \lambda_f, \bs \theta_c)$;
    \item solving the coarse model to obtain $\bs u_c(\bs \lambda_c)$;
    \item drawing a sample $\bs u_f \sim p_{cf}(\bs u_f|\bs u_c(\bs \lambda_c), \bs \theta_{cf})$.
\end{itemize}
In the examples presented in Section \ref{sec:examples},  we use the approximate posterior \eqref{LaplaceApprox} for $\tilde{\bs \theta}_c$ and  MAP estimates for all other  model parameters, which are denoted with a `$\tm{MAP}$' subscript in the following.
Since $p_{cf}(\bs u_f|\bs u_c(\bs \lambda_c), \bs \theta_{cf})$ is Gaussian, we can estimate the predictive posterior mean $\bs \mu_{\tm{pred}} = \left<\bu_f\right>_{p_{\tm{pred}}}$ as 
\begin{equation}
\bs \mu_{\tm{pred}}(\bsl_f) = \frac{1}{M} \sum_{m = 1}^M \int \bs{u}_f ~p_{cf}(\bs u_f| \bs u_c(\bs \lambda_c^{(m)}), \bs \theta_{cf}) d\bs u_f = \bs W_{\tm{MAP}} \frac{1}{M} \sum_{m = 1}^M\bs u_c(\bs \lambda_c^{(m)}) + \bs b_{\tm{MAP}}
\label{eq:mupred}
\end{equation}
where $\bs \lambda_c^{(m)} \sim p_c(\bs \lambda_c| \bs \lambda_f, \bs \theta_c) p(\bs \theta_c| \mathcal D)$. 
\revone{In the equation above, $M$ denotes the number of Monte Carlo samples needed to produce an accurate estimate of this quantity. As each of these samples requires solely  a solution of the coarse FE model, the cost is negligible. }
Similarly, the predictive posterior variance $\sigma_{\tm{pred},i}^2$ of each component $u_{f,i}$ can be estimated as: 
\be
\sigma_{\tm{pred},i}^2(\bsl_f) = \frac{1}{M} \sum_{m = 1}^M \int ( u_{f,i} -  \mu_{\tm{pred},i})^2 p_{cf}(\bs u_f| \bs u_c(\bs \lambda_c^{(m)}), \bs \theta_{cf}) d\bs u_f. 
\label{eq:sigmapred}
\ee

\subsubsection{Model testing}
In order to  assess the predictive performance of the model in the ensuing examples, we introduce the  following error measures
\begin{align}
\label{e_measure}
e (\bsl_f) &= \frac{1}{N_{dof,f}} \sum_{i = 1}^{N_{dof,f}} \frac{(\mu_{\tm{pred}, i}(\bsl_f) - u_{f, i}(\bsl_f) )^2}{\tm{var}(u_{f, i})}, \\
 L(\bsl_f)  &= - \frac{1}{N_{dof,f}}\sum_{i = 1}^{N_{dof,f}}\log \mathcal N(u_{f,i}^{(n)}| ~\mu_{\tm{pred}, i}(\bsl_f), \sigma_{\tm{pred}, i}^2(\bsl_f)),
\label{L_measure}
\end{align}
where the $u_{f, i}(\bsl_f)$ are the true FOM outputs. The error measure $e(\bsl_f)$ is normalized by the true variance $\tm{var}(u_{f,i})$ of $u_{f, i}$ (estimated by Monte Carlo). Hence, if we would naively use the training data mean as the predictive mean estimate for all test cases, the expected value $\left<e(\bsl_f)\right>$ would be 1. The second quantity $L(\bsl_f)$ represents an approximate predictive log-likelihood under the assumption that the predictive density can be sufficiently approximated by independent Gaussians. In contrast to $e(\bsl_f)$ which captures the deviation of the predictive mean from the truth, $L(\bsl_f)$ reflects also the predictive uncertainty.  To obtain a reference value for $ L(\bsl_f) $, we use the  means $\mu_{\tm{data},i}$ and variances $\sigma_{\tm{data},i}^2$ of the  training  data in place of  $\mu_{\tm{pred}, i}$ and $\sigma_{\tm{pred}, i}^2$ respectively. 
In the ensuing examples we compute average values of the aforementioned error measures over multiple samples $\bsl_f$ generated from the same density as the training data. 



\begin{figure}[t]
\centering
\begin{tikzpicture}[auto,node distance=0cm,>=stealth']
\footnotesize
\tikzset{block/.style= {draw, rectangle, rounded corners, minimum height=2em,minimum width=4em}}

\begin{pgfonlayer}{foreground}
\node[] (titleOffline) {Training/offline stage};
\node[block, node distance=1.6cm, fill=white, align=center, below of = titleOffline] (genData)
{Generate training data $\mathcal D$ \\
$\begin{aligned}
\bs \lambda_f^{(n)} &\sim p(\bs \lambda_f) \\
\bs u_f^{(n)} &= \bs u_f(\bs \lambda_f^{(n)}) \\
\mathcal D &= \left\{\bs \lambda_f^{(n)}, \bs u_f^{(n)} \right\}_{n = 1}^N
\end{aligned}$
};

\node[block, below of=genData, node distance=2cm, fill=white, align=center] (evalphiTrain)
{Evaluate features \\ $\varphi_{jk}(\bs \lambda_f^{(n)})$, sec. \ref{sec:p_c}};

\draw[->]     (genData) -- (evalphiTrain);

\node[block, below of=evalphiTrain, node distance=1.1cm, fill=white] (trainModel)
{Train ROM, sec. \ref{sec:modelTraining}};

\draw[->]     (evalphiTrain) -- (trainModel);

\node[block, below of=trainModel, node distance=10mm, fill=white, align=left] (MAP)
{Output: model posterior $p(\bs \theta| \mathcal D)$};

\draw[->]     (trainModel) -- (MAP);

\node[right of=MAP, node distance=30mm] (refPointA) {};

\draw[-]     (MAP) -- (refPointA.center);

\node[right of = titleOffline, node distance=65mm] (titleOnline) {\qquad\qquad\quad Prediction/online stage};

\node[block, below of=titleOnline, xshift=0mm, node distance=11mm, align=center, fill=white] (sample_theta)
{Sample model \\
parameters \\
$\bs \theta^{(s)} \sim p(\bs \theta| \mathcal D)$, eq. \eqref{LaplaceApprox}};

\node[left of=sample_theta, node distance=35mm] (refPointB) {};

\draw[-]    (refPointA.center) -- (refPointB.center);

\draw[->]    (refPointB.center) -- (sample_theta);

\node[block, right of =sample_theta, node distance=32mm, fill=white, align=center] (evalphi)
{Evaluate features \\ $\varphi_{jk}(\bs \lambda_{f, \tm{new}})$};

\node[block, right of=evalphi, node distance=26mm, align=center] (unseenLambda)
{Unseen \\ $\bs \lambda_{f, \tm{new}}$};

\draw[->]   (unseenLambda) -- (evalphi);

\node[block, below of =sample_theta, xshift=18mm, node distance=13mm, fill=white, align=center] (sample_lambda)
{Sample $\bs \lambda_c^{(m)} \sim p_c(\bs \lambda_c| \bs \lambda_{f, \tm{new}}, \bs \theta_{c}^{(s)})$};

\node[below of=sample_theta, node distance=11mm] (refPointC) {};

\node[below of=evalphi, node distance=11mm] (refPointD) {};

\draw[->]   (sample_theta) -- (refPointC);

\draw[->]   (evalphi) -- (refPointD);

\node[block, below of =sample_lambda, xshift=0mm, node distance=11mm, fill=white, align=center] (coarseSolver)
{Solve coarse model \\
$\bs u_c^{(m)} = \bs u_c(\bs \lambda_c^{(m)})$};

\draw[->]   (sample_lambda) -- (coarseSolver);

\node[block, below of =coarseSolver, xshift=0mm, node distance=12mm, fill=white, align=center] (reconstruction)
{Sample FOM solution \\
$\bs u_f^{(m)} \sim p_{cf}(\bs u_f|\bs u_c(\bs \lambda_c^{(m)}), \bs \theta_{cf})$};

\draw[->]   (coarseSolver) -- (reconstruction);
\node[below of=reconstruction, node distance=9mm] (refPointE) {};

\draw[-]    (reconstruction) -- (refPointE.center);

\node[block, left of =coarseSolver, xshift=0mm, node distance=38mm, align=center, fill=white] (repeat)
{Repeat and \\ update \\ $\bs \mu_{\tm{pred}}, \bs \sigma_{\tm{pred}}$ \\ eq. \eqref{eq:mupred}, \eqref{eq:sigmapred}};

\node[left of=refPointE, node distance=38mm] (refPointF) {};

\draw[-]    (refPointE.center) -- (refPointF.center);
\draw[-]    (refPointF.center) -- (repeat);

\node[above of=repeat, node distance=20mm] (refPointG) {};

\draw[-]    (repeat) -- (refPointG.center);

\node[right of=refPointG, node distance=2.5mm] (refPointH) {};

\draw[->]    (refPointG.center) -- (refPointH.center);
\end{pgfonlayer}{foreground}

\begin{pgfonlayer}{background}
\node[above of = evalphi] (outerOnline) {};
\node[fit= (outerOnline) (evalphi) (sample_theta) (sample_lambda) (coarseSolver) (reconstruction) (repeat) (refPointE), fill=lightgray!30, dashed, draw, inner sep=0.25cm, rounded corners] (Box)
{};
\node[above of =genData] (outerOffline) {};
\node[fit= (outerOffline) (genData) (evalphiTrain) (trainModel) (MAP), fill=lightgray!30, dashed, draw, inner sep=0.25cm, rounded corners] (Box)            {};
\end{pgfonlayer}{background}

\end{tikzpicture}
\caption{\revone{Model workflow for the training phase (left) and the prediction phase (right). 
}}
\label{offline_online_chart}
\end{figure}
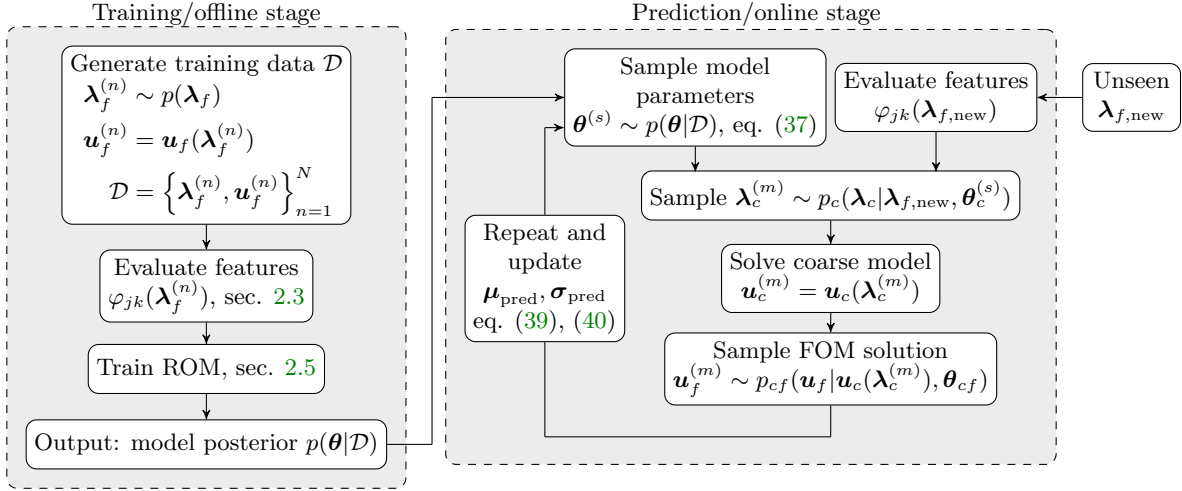

\revone{
\subsection{Numerical complexity analysis}
\label{sec:complexity}
In the complexity analysis of the proposed approach, it is essential to distinguish between training  and prediction  (Figure \ref{offline_online_chart}).
As it can be seen in the inner \tbf{for}-loop of Algorithm \ref{posteriorOpt}, training complexity grows linearly with the number of training samples $N$ due to the variational densities $q_n$ associated with each data point. However, as a result of the factorial form of the likelihood function in equation \eqref{likelihood1} and the resulting mutual independence of the $q_n$'s, this step may be fully parallelized in $N$. We did not observe any dependence of the required number of EM epochs w.r.t. $N$. 
}

\revone{
For training and prediction purposes, one must solve the coarse FE model. The cost of each of these solves depends on the dimension of $\bs{u}_c$ i.e. $N_{dof, c} = \dim(\bs u_c)$ which is by construction much smaller than that of $N_{dof, f} = \dim(\bs u_f)$.
Also, prediction complexity is completely independent of the number of training data $N$.
%
}

\revone{The scaling w.r.t. $N_{dof, f} = \dim(\bs u_f) \approx \dim(\bs \lambda_f)$ depends on the particular form of $p_{cf}$. For the one adopted in this study (Equation \eqref{pcf}) the scaling of the update equations in the training phase is linear with respect to $N_{dof, f}$. We note also that the values of the feature functions $\varphi(\bs \lambda_f)$ can be pre-computed and stored for each of the  training samples $ \lambda_f^{(n)}$. 
}

\section{Numerical experiments}
\label{sec:examples}
As a numerical test case for the method presented in the previous section, we consider the following linear elliptic PDE
\begin{align}
    \nabla_{\bs x} \cdot \left(-\lambda(\bx) \nabla_{\bs x} u \right) &= 0  & &\tm{for} \quad \bs x \in D = \left[0, 1\right]^d, \nonumber \\
        \label{poissonEq}
    u &= \hat{u} & &\tm{for} \quad \bs x \in \Gamma_u, \\
    -\lambda (\bx)(\nabla_{\bs x} u) \cdot \bs n &= \hat{\bs h} \cdot \bs n & &\tm{for} \quad \bs x \in \Gamma_{\bs h}, \nonumber
\end{align}
where \revone{$\bs n$ is the outward,  unit normal vector  on $\Gamma_{\bs h}$}, $\lambda(\bs x) > 0$ is a random diffusivity and $u = u(\bs x, \lambda(\bs x))$ the  solution field. 
The primary goal of the first example is to demonstrate the ability of the proposed model to identify salient, predictive features of the  random input $\bsl_f$. In the second example, the capability of the model to deal with very high-dimensional inputs (the cases considered involve $\tm{dim}(\bsl_f) = 256\times 256 = 65536$ and $\tm{dim}(\bsl_c)\le 64$) is evidenced as well as its resilience in providing accurate predictive estimates  with  limited training data ($N \approx 10\ldots 100$) or in cases where predictions are sought under different boundary conditions than the ones used in  training.

\subsection{One-dimensional example}
\label{sec:1d}
In the first example, we consider the SPDE  in  \eqref{poissonEq} in one spatial dimension $d = 1$ where there exists a closed-form solution for homogenized diffusion coefficients $\bs \lambda_c$. We use this closed-form solution as a feature function $\varphi(\bs \lambda_f)$ in combination with 99 other functions, some of which provide similar information.

We use the boundary conditions $\hat u(x) = 0$ for $x \in \Gamma_u = \{0\}$ and $\hat h = -100$ for $x \in \Gamma_{\bs h} = \{1\}$. 
The FOM is given by a Galerkin discretization with $N_{el, f} = 128$  linear  finite elements (i.e. $\tm{dim}(\bs{u}_f)=129$). 
In each such element, we assume constant diffusivity $\lambda_{f, i} \in \left\{\lambda_{lo}, \lambda_{hi} \right\}$, where  $\lambda_{lo} = 1$, $\lambda_{hi} = 10$. Samples of $\bs \lambda_f$ are generated by using a level-cut Gaussian process
\begin{equation}
f(x) \sim GP(0, k(x - x'))
\end{equation}
with  squared exponential covariance kernel $k(x - x') = \exp\left\{- \frac{(x - x')^2}{l^2} \right\}$  and  length scale parameter $l = 0.01$. We consider the values of $f(x)$ at the center points of each element which constitute a 128-dimensional Gaussian random vector $\bs f$. For each element $i$, we assign the value $\lambda_{lo}=1$ if $f_i  < f_{cut}$ and $\lambda_{hi}=10$ otherwise. 
 The cutoff parameter $f_{cut}$ is related to the expected volume fraction  of the two phases. 
For each training datum $\bsl_f^{(n)}$, we also randomize $f_{cut}$ such that the resulting dataset contains volume fractions uniformly distributed in $(0,1)$.

\subsubsection{The coarse-graining distribution \texorpdfstring{$p_c$}{}}
\label{sec:p_c1d}
%
For the coarse model, we employ a discretization consisting of 8 linear elements with the same boundary conditions as the FOM and assume that the diffusivity is constant within each element. Hence, $\tm{dim}(\bsl_c)=8$ and $\tm{dim}(\bu_c)=9$.
For the coarse-graining distribution $p_c(\bs \lambda_c|\bs \lambda_f, \bs \theta_c)$, we adopt the model\footnote{Since $\bs \lambda_f$ is bounded by $\lambda_{lo}, \lambda_{hi}$, we seek $\lambda_{c,k}$ that also take values in $[\lambda_{lo},  \lambda_{hi}]$. To enforce this constraint, we apply the sigmoid link function $\lambda_{c, k} = \chi(z_k) = \frac{\lambda_{hi} - \lambda_{lo}}{1 + e^{-z_k}} + \lambda_{lo}$ and perform the linear regression in $\bs z$-space. We note that more rigorous bounds \cite{Hashin1963, Torquato2001} exist in homogenization theory but are not applied here.} discussed in Section \ref{sec:p_c},
\begin{equation}
\lambda_{c,k} = \sum_{j = 1}^{N_{\tm{features}}} \tilde{\theta}_{c, j} \varphi_j(\bs \lambda_f^{[k]}) + \sigma_{c,k} Z_k, \qquad Z_k \sim \mathcal N(0, 1),
\end{equation}
where with $\bs \lambda_f^{[k]}$ we denote the subset of $\bs \lambda_f$ which is part of coarse element $k$ i.e. for the first coarse element $\lambda_f^{[k]}$ corresponds to the first 8 entries of $\bsl_f$ and so on. 
We employ the same  feature functions in all coarse elements, i.e. $\varphi_{jk} = \varphi_{j}$. Furthermore, we assume that the same coefficients can be used in each of those regressions, i.e. $\tilde{\theta}_{c,jk} = \tilde{\theta}_{c,j}$.  As a result, we obtain closed-form updates for the model parameters $\bt_c$ which, according to Equations \eqref{grad_thetac}, \eqref{dF_dSigma} will take the form
\begin{small}
\begin{align}
\label{thetaUpdate}
\tilde{\bs \theta}_c^{(t + 1)} &= \left(\sum_{n = 1}^N  \bs \Phi^T(\bs \lambda_f^{n})(\bs \Sigma_c^{(t)})^{-1}\bs \Phi(\bs \lambda_f^{(n)}) + (\tm{diag}[\bs \gamma^{(t)}])^{-1}\right)^{-1} \sum_{n' = 1}^N \bs \Phi^T(\bs \lambda_f^{(n')})(\bs \Sigma_c^{(t)})^{-1} \left<\bs z^{(n')} \right>_{q_{n'}^{(t + 1)}}, \\
\bs \Sigma_c^{(t + 1)} &= \frac{1}{N} \sum_{n = 1}^N\left<\tm{diag}[(\bs z^{(n)} - \bs \Phi(\bs \lambda_f^{(n)})\tilde{\bs \theta}_c)(\bs z^{(n)} - \bs \Phi(\bs \lambda_f^{(n)})\tilde{\bs \theta}_c)^T] \right>_{q_n^{(t + 1)}},
\label{sigmaUpdate}
\end{align}
\end{small}
where $\Phi_{kj}(\bs \lambda_f) = \varphi_j(\bs \lambda_f^{[k]})$ and $\bs \gamma$ can be updated according to \eqref{updateGamma}. We assume that $p(\bs \Sigma_c|\mathcal D) = \delta(\bs \Sigma_c - \bs \Sigma_{c,\tm{MAP}})$ and that $p(\tilde{\bs \theta}_c|\mathcal D)$ is given by the Laplace approximation in \eqref{LaplaceApprox}.

\paragraph{Feature functions}
It is known \cite{Torquato2001} that the effective diffusion coefficient  for 1-dimensional problems such as the one considered, corresponds to  the harmonic mean.   We therefore use it as a  feature function in conjunction with 99 other ones which include generalized means, lineal path function \cite{Lu1992},  2-point correlation function, effective medium approximations \cite{Torquato2001} and distance transforms \cite{Rosenfeld1966, Maurer2003}.

\subsubsection{The coarse-to-fine map \texorpdfstring{$p_{cf}$}{}}
For the coarse-to-fine map $p_{cf}(\bs u_f| \bs u_c(\bs \lambda_c), \bs \theta_{cf})$, we employ the model given in \refeq{pcf} and set the bias parameter $\bs b = 0$. We further determine the projection matrix $\bs W \in \RR^{129 \times 9}$ by linearly interpolating between coarse and fine grids. In particular 
\begin{equation}
    W_{ij} =
    \begin{cases}
    &\frac{x_i - X_{j - 1}}{X_j - X_{j - 1}} \qquad \tm{for} \qquad X_{j - 1} \le x_i \le X_j, \\
    &\frac{x_i - X_{j + 1}}{X_j - X_{j + 1}} \qquad \tm{for} \qquad X_{j} \le x_i \le X_{j+1}, \\
    &0 \qquad \qquad ~~ \tm{else},
    \end{cases}
\end{equation}
where $X_j=\frac{j-1}{8}, j=1,\ldots,\tm{dim}(\bu_c)=9$ are the coordinates of the nodes of the coarse model (i.e. the spatial locations to which  the outputs $\bu_c$  correspond to) and $x_i=\frac{i-1}{128}, i=1,\ldots,\tm{dim}(\bu_f)=129$ are the coordinates of the nodes of the FOM (i.e. the spatial locations to which  the outputs $\bu_f$  correspond to).
The covariance matrix $\bs S$ (\refeq{pcf}, which is assumed to be diagonal) is treated as  free parameter and its MAP estimate is computed. In the absence of a prior, according to \eqref{grad_S}, the updates for $\bs S$ are closed-form,
\begin{equation}
\bs S^{(t + 1)} = \frac{1}{N}\sum_{n = 1}^N \left< \tm{diag}\left[(\bs u_f^{(n)} - \bs W \bs u_c(\bs \lambda_c^{(n)}))(\bs u_f^{(n)} - \bs W \bs u_c(\bs \lambda_c^{(n)}))^T\right] \right>_{q_n^{(t + 1)}}.
\label{S_update}
\end{equation}

\begin{figure}[t]
  \centering
  \includegraphics[width=\textwidth]{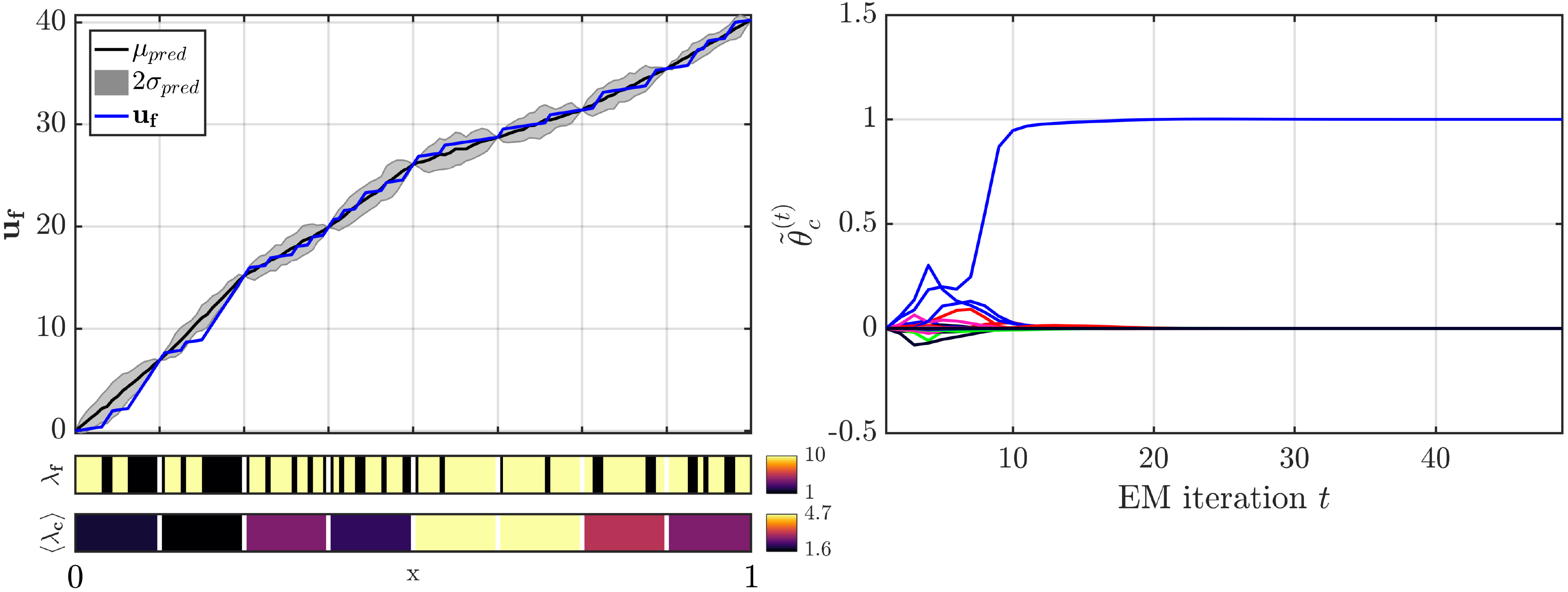}
  \caption{Left: One-dimensional example. For a test input $\bsl_f$, the blue line corresponds to true FOM output $\bs u_f$, the black line  is the predictive mean $\bs{\mu}_{\tm{pred}}$ (\refeq{eq:mupred}) enveloped by $\pm 2$ predictive standard deviations $\sigma_{\tm{pred}}$ (\refeq{eq:sigmapred}). The  bars underneath depict the FOM input $\bsl_f$ ($\tm{dim}(\bsl_f)=128$) (top) and the predictive posterior mean $\left<\bs \lambda_c \right>_{p_c}$ with the model parameters learned from the training data. (Right) Evolution of $\tilde{\bs \theta}_c^{(t)}$ with respect to  EM iterations $t$. The blue curve corresponds to the harmonic-mean feature function and quickly converges to $1$. All remaining 99 coefficients become $0$ (only a subset is depicted).
}
  \label{fig:1d_c=10_pred}
\end{figure}

\subsubsection{Results}
In Figure \ref{fig:1d_c=10_pred} results obtained with $N=16$ training data are depicted. On the right-hand side, we observe the evolution of the coefficients  $\tilde{\bs \theta}_c$ with respect to the Expectation-Maximization iterations. One observes that  $\tilde\theta_{c,1} $, which corresponds to the harmonic mean feature function, quickly converges to $1$ whereas all remaining $\tilde{\bs \theta}_c$'s become $0$, i.e. all remaining features are deactivated. Hence the sparsity-inducing prior is shown  capable of distinguishing the most predictive feature function(s), despite the large number of such features and the small number of training data. 
On the left hand-side, we depict predictions of the FOM output $\bu_f$ obtained using the trained model for an indicative test case $\bsl_f$. While the posterior mean does not coincide with the reference solution, the model's predictive posterior is able to envelop it. One can also visually inspect the predictive  posterior means of the coarse model properties $\bsl_c$ in relation with the underlying FOM diffusivity $\bsl_f$.

In order to assess the overall predictive ability of the model we computed average values of the error metric $e(\bsl_f)$ (\refeq{e_measure}) in Section \ref{sec:modelpred} over $N_{\tm{test}} = 1024$ test samples. We obtain the value of $\left<e(\bsl_f)\right>= 0.027(3)$  which is approximately 30 times smaller than the  reference value of $1$.

\subsection{Two-dimensional examples}
\label{2dexamples}
\begin{figure}[t]
  \centering
  \includegraphics[width=.95\textwidth]{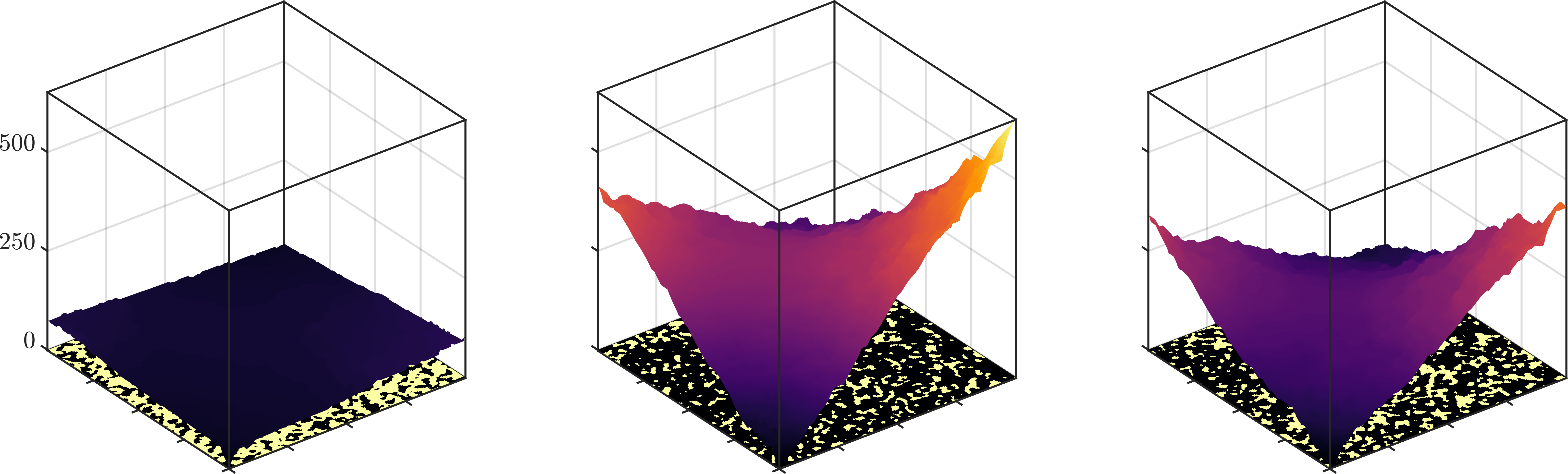}
  \caption{Two-dimensional example. Samples $\bs \lambda_f$, with $\tm{dim}(\bs \lambda_f) = 256\times 256$ and the corresponding PDE outputs $\bs u_f$ ($\tm{dim}(\bu_f)=257 \times 257$) for contrast $\frac{\lambda_{hi}}{\lambda_{lo}} = 100$. The same  boundary conditions are employed. 
}
  \label{microstructures}
\end{figure}
In this section, we examine the SPDE in \eqref{poissonEq} in the two dimensional unit square where there is no closed form solution for the effective diffusion coefficients $\bs \lambda_c$. For the FOM, we discretize with a uniform square mesh of size $256\times 256$ (\textbf{$\tm{dim}(\bu_f)=66049$}) and assume constant diffusivity within each element (i.e. \textbf{ $\tm{dim}(\bsl_f)=65536$}). We consider two-phase random media, i.e.  $\lambda_{f,i} \in \left\{ \lambda_{lo}, \lambda_{hi}\right\}$ and evaluate the performance of the method proposed for various contrasts $c=\frac{\lambda_{hi}}{\lambda_{lo}}$\footnote{\revone{We always used  $\lambda_{lo}=1$ and set $\lambda_{hi}=c$. While the coercivity constant of the PDE depends on  $\lambda_{lo}$, the data-driven model proposed was found to be  insensitive to this.}  }. It is noted that the more pronounced the contrast in the properties of the two phases is, the more the (random) topology and higher-order statistical descriptors affect the macroscopic response (\cite{Torquato2001}).
We consider a distribution on $\bsl_f$ defined implicitly through a level-cut Gaussian process
 $f(\bs x) \sim GP(0, k(\bs x - \bs x'))$ with $k(\bs x - \bs x') = \exp \left\{-\frac{|\bs x - \bs x'|^2}{l^2} \right\}$ and $l = 0.01$. We generate samples of the random vector associated with the center points of each of the $65536$ elements (e.g. \cite{shinozuka_simulation_1996}) and assign values $ \lambda_{lo}$ or $\lambda_{hi}$ based on a threshold $f_{cut}$, as we did in Section \ref{sec:1d}. We again randomize this threshold so as the resulting samples have a range of expected volume fractions between $0$ and $1$. 
Indicative samples $\bs \lambda_f$ are depicted in Figure \ref{microstructures} together with the corresponding FOM outputs $\bu_f(\bsl_f)$.

We consider  boundary conditions of the form
\begin{align}
\nonumber
\hat u(\bs x) &= a_0 + a_1 x_1 + a_2 x_2 + a_3 x_1 x_2,  &\bs x \in \Gamma_u, \\ 
    \hat{\bs h}(\bs x) &= -\nabla_{\bs x} \hat u(\bs x),  &\bs x \in \Gamma_{\bs h}.
\label{boundaryConditions}
\end{align}
Furthermore, we use $\Gamma_u = \left\{\bs 0 \right\}$ and $\Gamma_{\bs h} = \pa D\backslash \left\{\bs 0 \right\}$, i.e. Neumann boundary conditions of the form above almost everywhere.

\subsubsection{Model distributions}

For the coarse-to-fine map $p_{cf}(\bs u_f|\bs u_c(\bs \lambda_c), \bs \theta_{cf})$, we again fix the bias vector $\bs b = \bs 0$ and the coarse-to-fine projection matrix $\bs{W}\in \RR^{66049 \times \tm{dim}(\bu_c)}$  so that it corresponds to a bilinear interpolation of the fine and coarse model grid points (as we did in the one-dimensional example).  The covariance $\bs S$ of the residual noise in \refeq{pcf} is treated as  a free parameter and the MAP estimate is obtained using the  same updates  as in \refeq{S_update}.

For the coarse-graining distribution $p_c(\bs \lambda_c|\bs \lambda_f, \bs \theta_c)$, we use the relation
\begin{equation}
\lambda_{c,k} = \sum_{j = 1}^{N_{\tm{features}}} \tilde{\theta}_{c, j} \varphi_j(\bs \lambda_f^{[k]}) + \sigma_{c,k} Z_k, \qquad Z_k \sim \mathcal N(0, 1),
\label{eq:ex2pc1}
\end{equation}
(as in \ref{sec:p_c1d}) with a set of 100 feature functions adapted to the $2d$ case (see Appendix \ref{sec:featureList} for a summary). We employed the same coefficients $\tilde{\theta}_{c, j}$ for all macro-elements $k$ and will discuss a more flexible version in Section \ref{sec:ex2flex}. 
The update equations for $\tilde{\bs \theta}_c, \bs \Sigma_c$ are equivalent to those given in \eqref{thetaUpdate}, \eqref{sigmaUpdate} and we employ  $p(\bs \Sigma_c|\mathcal D) = \delta(\bs \Sigma_c - \bs \Sigma_{c,\tm{MAP}})$ and $p(\tilde{\bs \theta}_c|\mathcal D)$ as computed  by the Laplace approximation in \eqref{LaplaceApprox}.

\subsubsection{Predictive performance}
\begin{figure}[t]
  \centering
  \includegraphics[width=.95\textwidth]{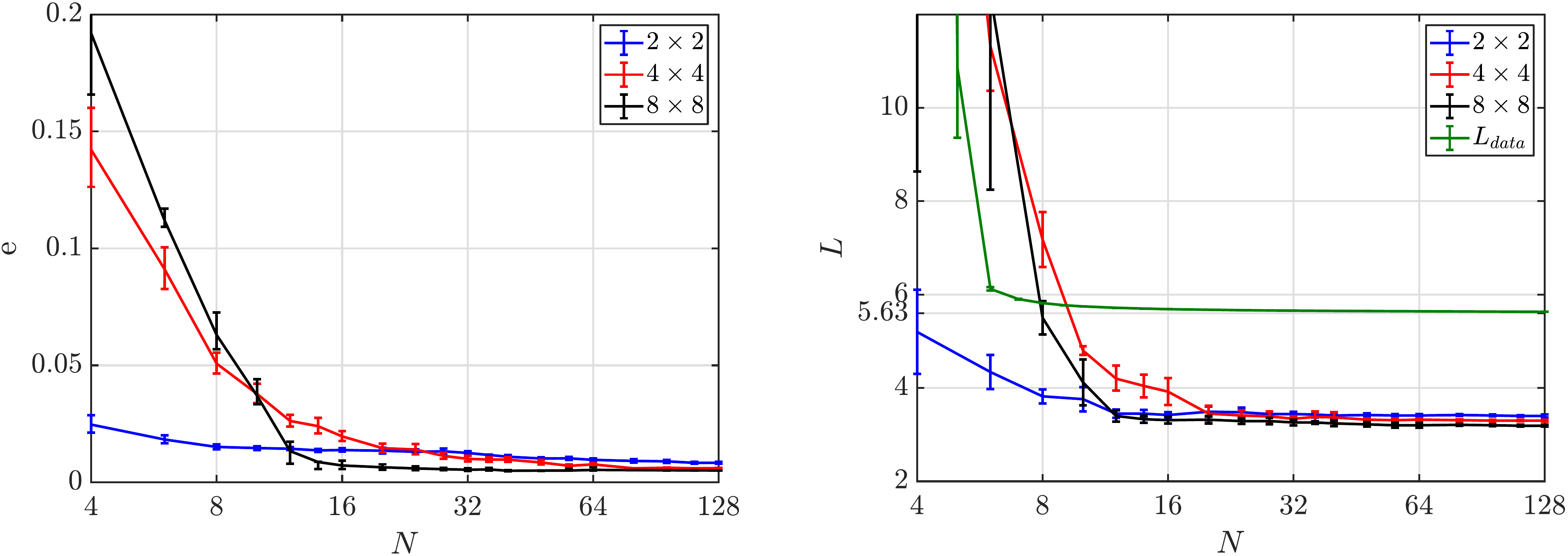}
  \caption{Averaged error measures $e$ and $ L $ as defined in equations \eqref{e_measure}, \eqref{L_measure} for contrast $c = \frac{\lambda_{hi}}{\lambda_{lo}} = 2$ and different coarse model  sizes $N_{el, c}$ versus the number of training data samples $N$. The error bars are due to randomization of training data. The green line on the right corresponds to $L_{\tm{data}}$, see Section \ref{sec:modelpred}. 
  }
  \label{errorPlot}
\end{figure}
\begin{figure}[!t]
  \centering
  \includegraphics[width=.95\textwidth]{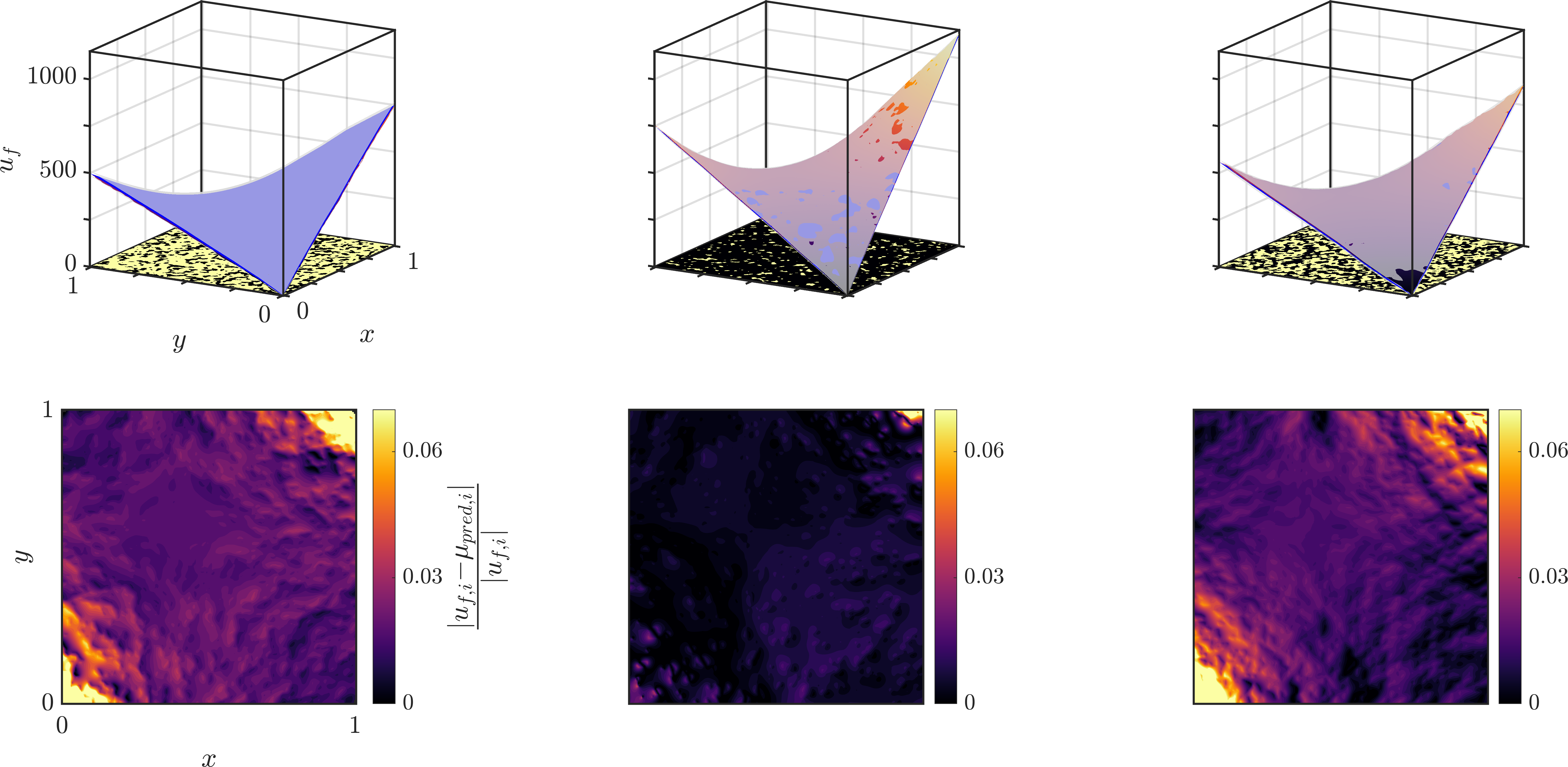}
  \caption{Top row: Predictive mean $\bs \mu_{\tm{pred}}$ (blue) $\pm \bs \sigma_{\tm{pred}}$ (transparent grey) and true response $\bs u_f$ (colored) for three test samples for $c = \frac{\lambda_{hi}}{\lambda_{lo}} = 2$, $N_{el,c} = 8\times 8$ and $N = 128$ training data samples. \revone{Bottom row: Normalized error between ground truth $\bs{u}_f$ and posterior predictive mean $\bs{\mu}_{pred}$ i.e.  $\frac{|u_{f,i} - \mu_{pred, i}|}{|u_{f,i}|}$.}}
  \label{predictions_c=2}
\end{figure}

In order to assess the predictive performance of the proposed model, we use the error measures $e$ and $L$ as defined in \eqref{e_measure} and \eqref{L_measure}, respectively and average over multiple test cases.
Both measures are plotted in Figure \ref{errorPlot}  against the number of training samples $N$ for the three different coarse model sizes with $N_{el, c} = 2\times 2, 4\times 4, 8\times 8$ and for a contrast  $c = \frac{\lambda_{hi}}{\lambda_{lo}} = 2$. The test data are generated with boundary conditions as in \refeq{boundaryConditions} with $\bs a = (0,~ 800,~ 1200,~ -2000)^T$. We observe that in all three cases, the reduced-order models constructed are able to reach their asymptotic values with less than $N=16$ training samples. The coarsest of these models (i.e. with $N_{el, c} = 2\times 2$) converges the fastest due to the fewer free parameters but attains error values that are not as low as the finer models. In the top row of Figure \ref{predictions_c=2}, three indicative test samples $\bs \lambda_f, \bs u_f$ are depicted and compared with the posterior predictive estimates $\bs{\mu}_{\tm{pred}}$ (\refeq{eq:mupred}) and $\sigma_{\tm{pred}}$ (\refeq{eq:sigmapred}), whereas the bottom row shows the $L^2$-distance of the predictive mean to the true reference. The latter are computed with $N=128$ training samples and for a coarse model of size  $N_{el,c} = 8\times 8$. We observe that in all cases  and despite the unavoidable predictive uncertainty, the probabilistic predictions obtained tightly envelop the truth. 

\begin{figure}[h!]
  \centering
  \begin{subfigure}[c]{0.76\textwidth}
  \centering
  \includegraphics[width=\textwidth]{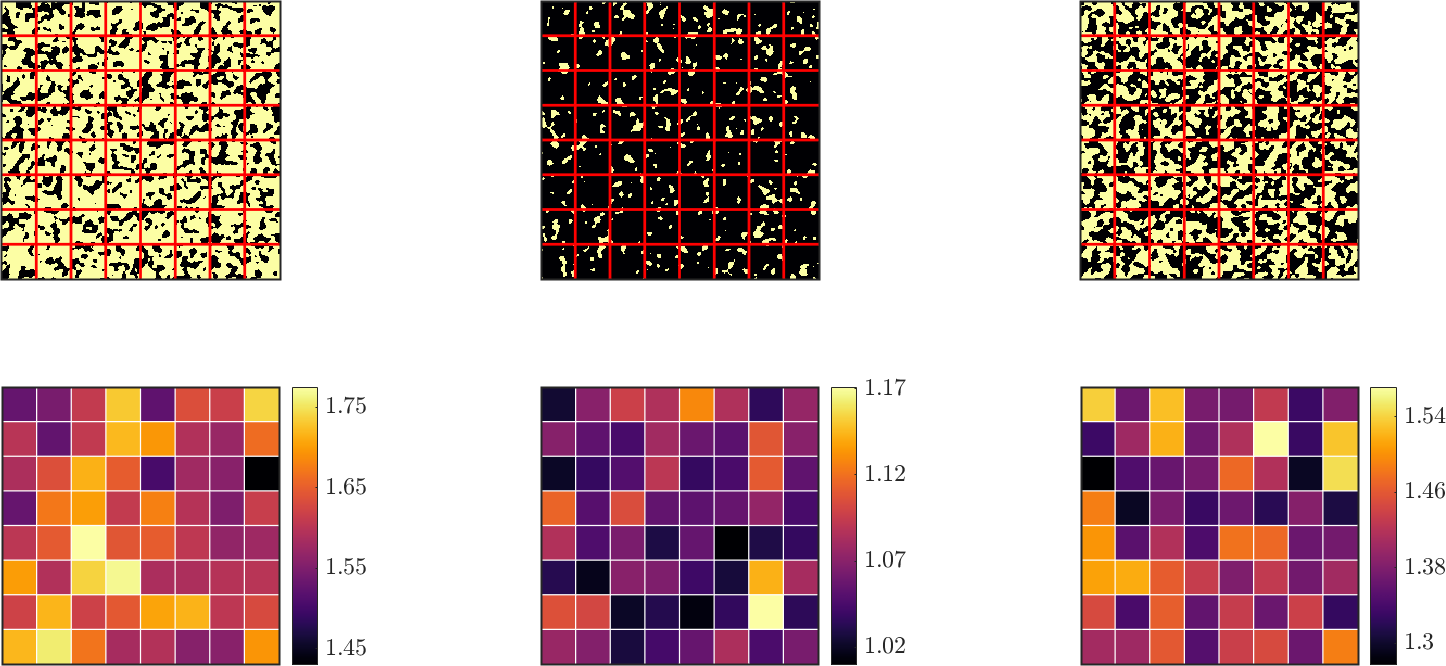}
  \caption{Coarse-grained, effective property $\left<\bs \lambda_c \right>_{p_c}$ for the three test samples shown in Figure \ref{predictions_c=2} with $N = 128$, $N_{el,c} = 8\times 8$.}
  \label{effPropa}
  \end{subfigure}
  \begin{subfigure}[c]{0.76\textwidth}
  \centering
  \includegraphics[width=\textwidth]{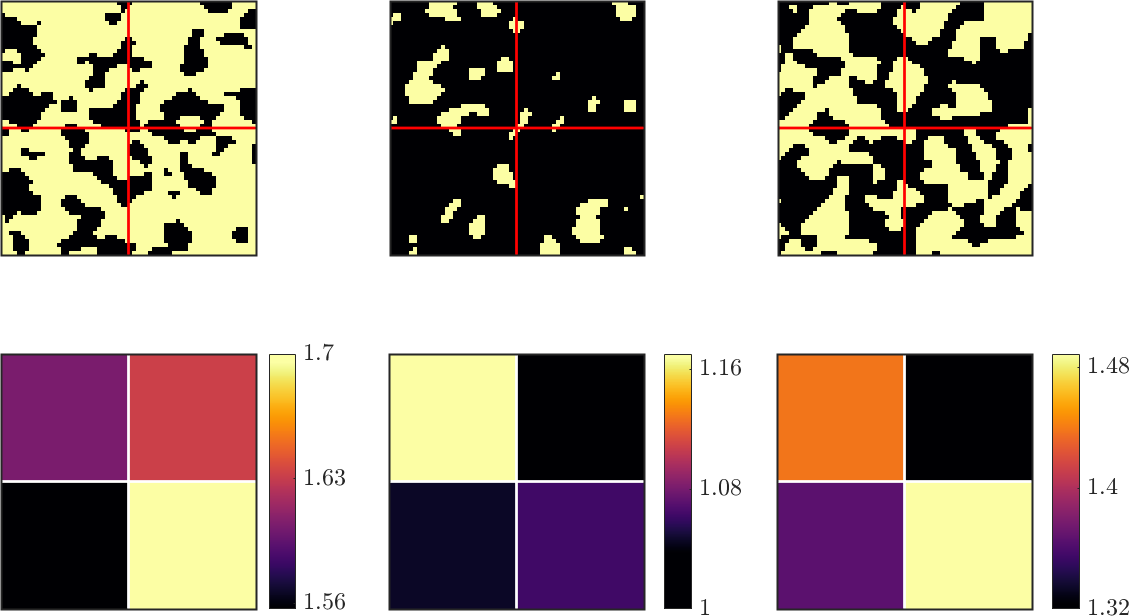}
  \caption{Lower right corner macro-cells and mean effective properties $\left<\lambda_{c,k}\right>$ of the microstructures shown in \ref{effPropa}}
  \end{subfigure}
  \begin{subfigure}[c]{0.55\textwidth}
  \centering
  \includegraphics[width=\textwidth]{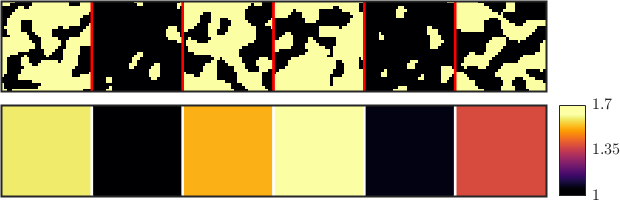}
  \caption{Effective properties of randomly chosen macro-cells of the microstructures shown in \ref{effPropa}.}
  \end{subfigure}
  \caption{Predictive posterior mean  $\left<\bs \lambda_c\right>$.}
  \label{effProp}
\end{figure}
\begin{figure}[t]
  \centering
  \includegraphics[width=.8\textwidth]{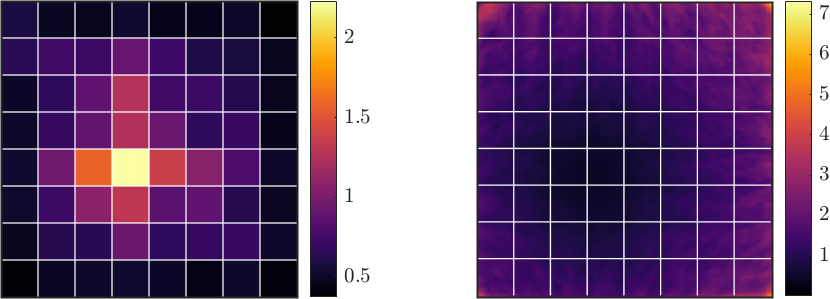}
  \caption{MAP estimates of $\sigma_{c,k}$ (left) and ${s_j}$ (right) as computed for $c = 2$, $N = 128$ and $N_{el,c} = 8\times 8$.}
  \label{predUncertainty}
\end{figure}

 Figure \ref{effProp}  provides  further insight  on the trained model as it depicts the corresponding predictive posterior means of the coarse-model's properties $\bsl_c$ for various test instances $\bsl_f$. 
The predictive uncertainty $ \sigma_{\tm{pred}}$ (\refeq{eq:sigmapred}) is in part due  to the residual uncertainty in $p_{cf}$ captured $\bs S=\tm{diag}(s_j^2)$ as well as the uncertainty in $p_c$ modeled by $\sigma_{c,k}^2$.   The corresponding standard deviations $\sigma_{c,k}, {s_j}$ for each of the coarse element $k$ and FOM nodes $j$ are depicted in Figure \ref{predUncertainty}. We observe that the $\sigma_{c, k}$ is  generally larger way from the boundary of the problem domain $D$. The opposite behavior is observed for the ${s_j}$'s  which tend to be larger closer to the boundaries. 

\subsubsection{Activated features for different contrasts}

\begin{figure}[t!]
  \centering
  \includegraphics[width=\textwidth,height=10cm]{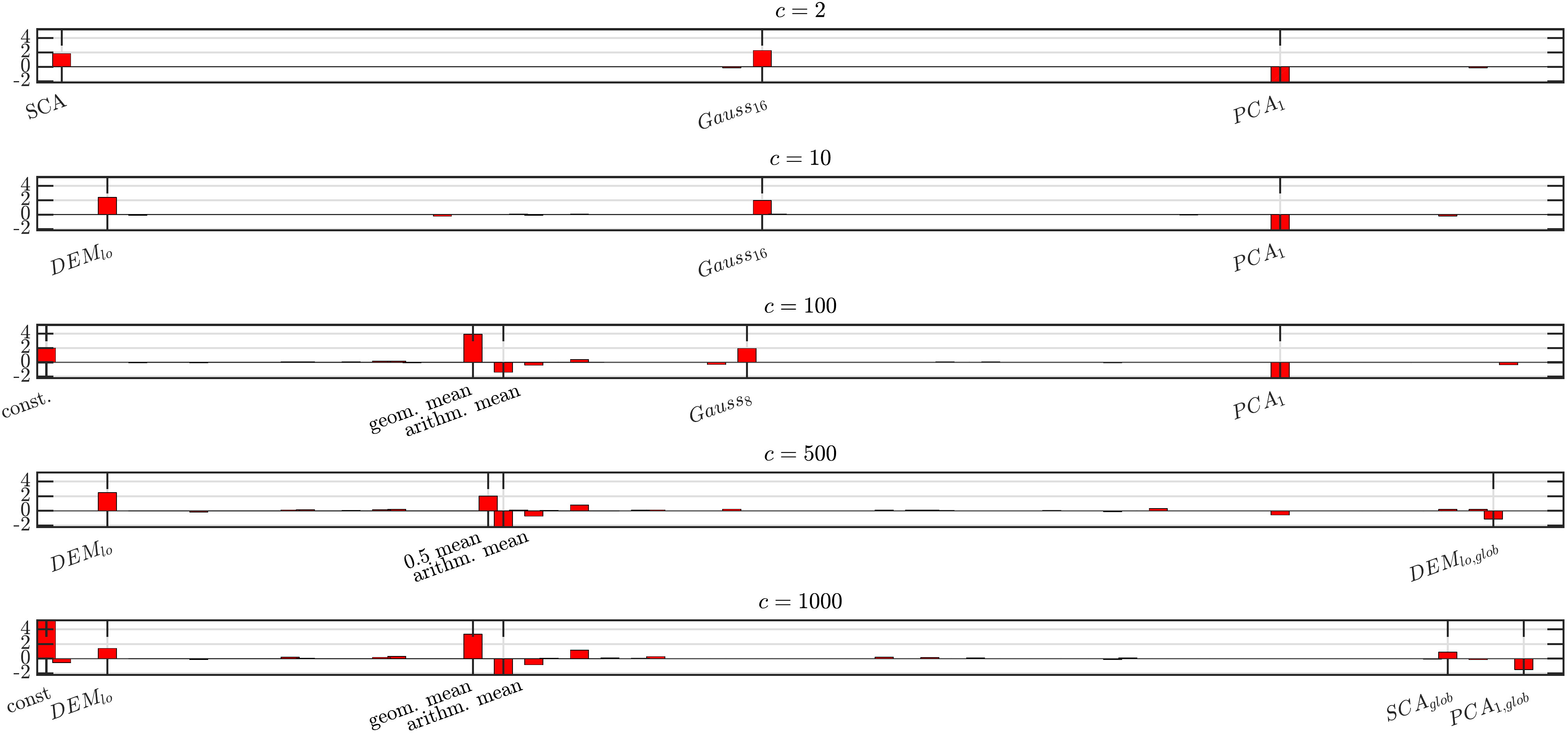}
  \caption{MAP estimates of the coefficients $\tilde{\bs \theta}_c$  for 5 different contrast ratios $c$. 
  }
  \label{activatedFeatures}
\end{figure}

\begin{figure}[h]
  \centering
  \includegraphics[width=.9\textwidth]{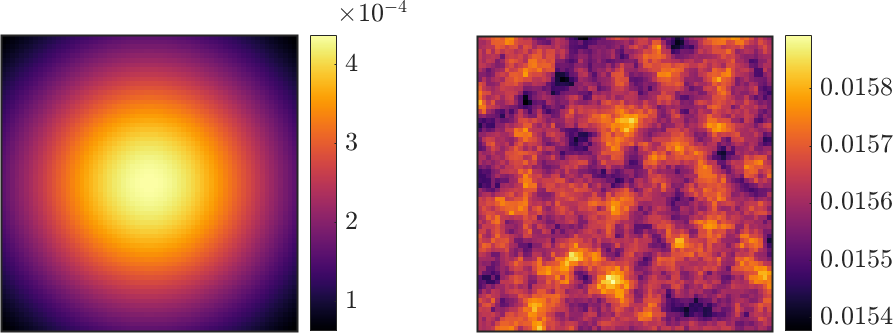}
  \caption{Left: The ``Gaussian linear filter'' with variance $\bs \Sigma = 8L\bs I$ (marked as $\tit{Gauss}_8$ in Figure \ref{activatedFeatures}) where $L$ is the length of the macro-cell. Right: The first PCA component (marked as $\tit{PCA}_1$ in \ref{activatedFeatures}) obtained performing PCA on all macro-cells of 4096 unsupervised samples of $\bs \lambda_f^{[k]}$. The feature function outputs are computed as the inner product of the above images with every $\bs \lambda_f^{[k]}$.}
  \label{linFilt}
\end{figure}

In order to gain further insight of the feature functions that are activated, we train the coarse model 
 of size $N_{el,c} = 4\times 4$ for five  different contrast values $c = 2, ~10, ~100, ~500$ and $1000$.  We generate $N = 1024$ in which we also randomize the boundary conditions employed by drawing $\bs a \sim \mathcal N(\bs 0, \bs \sigma_{\bs a}^2)$ in \refeq{boundaryConditions} with $\bs \sigma_{\bs a}^2 = \bsmat 0, & 10^6, & 10^6, & 10^6 \esmat^T$. The MAP estimates of the coefficients $\tilde{\bs \theta}_c$ are shown in figure \ref{activatedFeatures}. 
We generally observe that for higher contrast values  $c$, the magnitude  of the non-zero  $\tilde{\theta}_c$'s  as well as the number of activated feature functions increase. Furthermore  feature functions  taking into account the whole microstructural vector $\bs \lambda_f$ become activated. This could be attributed to the fact that the higher the contrast the more prominent becomes the role of the microstructure and its higher-order statistics in predicting the system's response.
Apart from generalized means, other  features that play a role correspond to  effective medium approximations  such as the self-consistent approximation (SCA) or Bruggeman formula \cite{Bruggeman1935} as well as the differential effective-medium approximation (DEM) \cite{Bruggeman1935}. Statistical features such as  ``Gaussian linear filters'' (Figure \ref{linFilt}) and the first principal component  (computed by PCA on 4096 samples of  $\bs \lambda_f$) also seem to be important. 
%

\subsubsection{Predictions  under different  boundary conditions}
\begin{table}[h]
\begin{small}
\begin{center}
\tbf{Error measure} $\left< e \right>$
\end{center}
\begin{center}
\begin{tabular}{l||c|c}
\backslashbox{predict}{trained on}
&\makebox[4cm]{$\bs a = \bsmat 0 & 800 & 1200& -2000\esmat^T$}&\makebox[4cm]{$\bs a' = \bsmat 0~ 500~ -1500~ 1000\esmat^T$}\\\hline\hline
$\bs a = \bsmat 0 & 800 & 1200& -2000\esmat^T$ & $0.00895$ & $0.00963$  \\\hline
$\bs a' = \bsmat 0~ 500~ -1500~ 1000\esmat^T$ & $0.0102$ & $0.00950$ 
\end{tabular}
\vspace{2mm}
\end{center}
\begin{center}
\tbf{Error measure} $\left<L \right>$
\end{center}
\begin{center}
\begin{tabular}{l||c|c}
\backslashbox{predict}{trained on}
&\makebox[4cm]{$\bs a = \bsmat 0 & 800 & 1200& -2000\esmat^T$}&\makebox[4cm]{$\bs a' = \bsmat 0~ 500~ -1500~ 1000\esmat^T$}\\\hline\hline
$\bs a = \bsmat 0 & 800 & 1200& -2000\esmat^T$ & $4.06$ & $4.31$  \\\hline
$\bs a' = \bsmat 0~ 500~ -1500~ 1000\esmat^T$ & $3.90$ & $3.86$ 
\end{tabular}
\end{center}
\caption{Averaged error measures $e$ and $ L $ as defined in \refeq{e_measure} and \refeq{L_measure}.  
In the off-diagonal cells, we test on data with boundary conditions $\bs a$ whilst having trained using $\bs a'$ and vice versa.}
\label{stableBCtable}
\end{small}
\end{table}
\begin{figure}[h]
  \centering
    \begin{subfigure}[c]{0.45\textwidth}
  \centering
  \includegraphics[width=\textwidth]{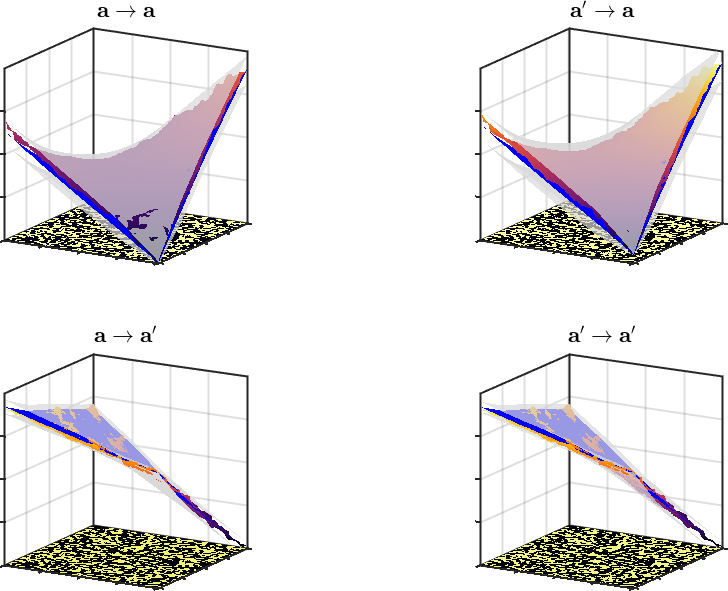}
  \caption{Predictive mean $\bs \mu_{\tm{pred}}$ (blue) $\pm \bs \sigma_{\tm{pred}}$ (transparent grey) and true response $\bs u_f$ (colored).  The top left and the bottom right plots are predictions using identical boundary conditions as have been used for training. The top right and the bottom left are trained using $\bs a'$ but predict on $\bs a$ and vice versa. 
  }
  \label{crossPred}
  \end{subfigure}
  \hfill
  \begin{subfigure}[c]{0.45\textwidth}
  \centering
  \includegraphics[width=\textwidth]{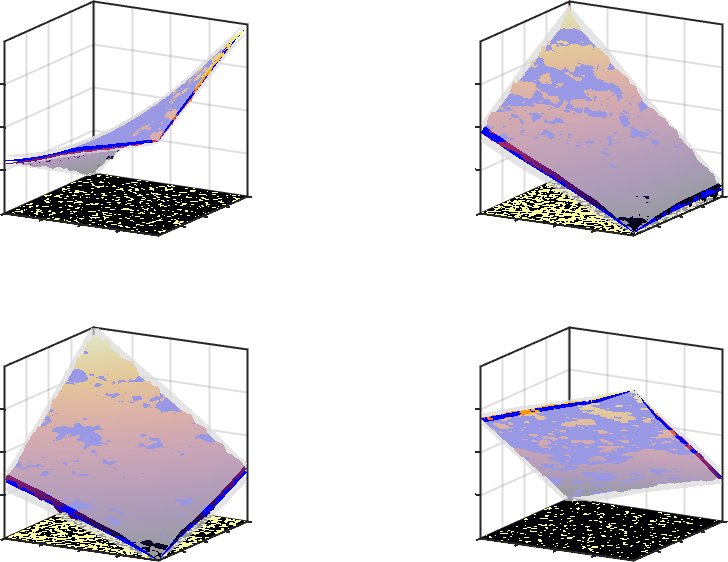}
  \caption{Predictive mean $\bs \mu_{\tm{pred}}$ (blue) $\pm \bs \sigma_{\tm{pred}}$ (transparent grey) and true response $\bs u_f$ (colored). The boundary conditions of the four test cases are randomly selected as explained in the text whereas training was performed using boundary conditions $\bs a$.}
  \label{predRandBC}
  \end{subfigure}
  \caption{
  Prediction examples for $N = 1024$, $N_{el,c}= 4\times 4$, $l = 0.01$ and $c = 10$ using different boundary conditions. No substantial deterioration in predictive performance is observed if predictions are performed on a test set with different boundary conditions than the training set.}
\end{figure}

The goal of this section is to examine the ability of the proposed model to produce accurate predictions of FOM outputs under certain boundary conditions when it has been trained with data involving FOM runs under different boundary conditions. 
%
%
%
To investigate this, we train the model with a coarse model size $N_{el, c} = 4\times 4$  and FOM data obtained under the two boundary conditions specified by (\refeq{boundaryConditions}) $\bs a = (0,~ 800,~ 1200,~ -2000)^T$ and $\bs a' = (0,~ 500,~ -1500,~ 1000)^T$. 
We use $N = 1024$ training samples in order to avoid the effects of small datasets. The predictive error measures  $e$ and $L$ (Section \ref{sec:modelpred}) are averaged  over multiple test instances and the results are shown in Table \ref{stableBCtable}.  We observe only slight deterioration for predictions on different boundary conditions than those used for training which implies that the model is able to incorporate salient information about the physical behavior of the random medium. In Figure \ref{crossPred} a few indicative test cases are depicted, one for each of the four possible combinations of training/testing boundary conditions.  In Figure \ref{predRandBC} we show 4 test cases where the model  is trained with FOM data obtained on boundary conditions  $\bs a$ and predictions are computed for  randomly sampled boundary conditions according to $\tilde{\bs a} \sim \mathcal N(\bs 0, \bs \sigma_{\bs a}^2)$ with $\bs \sigma_{\bs a}^2 = \bsmat 0, & 10^6, & 10^6, & 10^6 \esmat^T$. In all the aforementioned cases, accurate predictions were obtained which envelop the ground truth.

\subsubsection{Predictive performance improvement by local/global \texorpdfstring{$\tilde{\bs \theta}_c$}{}'s}
\label{sec:ex2flex}

\begin{table}[h]
\begin{small}
\begin{center}
\tbf{Error measure} $e$
\begin{tabular}{c||c|c||c|c} 
 & \multicolumn{1}{l}{$N = 16$} & & \multicolumn{2}{l}{$N = 1024$} \\
c &$\tilde{\theta}_{c,jk} = \tilde{\theta}_{c,j}$ & $\gamma_{jk} = \gamma_{jk'}$ &$\tilde{\theta}_{c,jk} = \tilde{\theta}_{c,j}$ & $\gamma_{jk} = \gamma_{j}$ \\\hline\hline
2 & $0.0215 \pm 0.002$ & $0.0715 \pm 0.036$ & 0.00643 & 0.00412 \\\hline
10 & $0.0277 \pm 0.027$ & $0.0479 \pm 0.009$ & 0.00948 & 0.00593 \\\hline
100 & $0.0317 \pm 0.005$ & $0.0589 \pm 0.015$ & 0.0166 & 0.0103
\end{tabular}
\vspace{2mm}
\end{center}
\end{small}
\caption{
Averaged error measure $e$ as defined in \eqref{e_measure} for a model with  $\tilde{\theta}_{c,jk} = \tilde{\theta}_{c,j}$ for all macro-cells $k$ (\refeq{eq:ex2pc1}) and for the model $\tilde{\theta}_{c, jk} \neq \tilde{\theta}_{c, jk'}$ (\refeq{eq:ex2pc2}) but identical  hyperparameters  $\gamma_{jk} = \gamma_j$. Predictions for $N=16$ and $N=1024$ training data are reported.
}
\label{nonLocalTable}
\end{table}
We consider in this section a more flexible model for $p_c$ and examine its potential in terms of the accuracy of the predictions produced. In contrast to \refeq{eq:ex2pc1}, we consider relations between $\bsl_c$ and $\bsl_f$ of the form
\begin{equation}
z_k = \sum_{j = 1}^{N_{\tm{features}}} \tilde{\theta}_{c, jk} \varphi_j(\bs \lambda_f^{[k]}) + \sigma_{c,k} Z_k, \qquad Z_k \sim \mathcal N(0, 1),
\label{eq:ex2pc2}
\end{equation}
where the coefficients $\tilde{\theta}_{c, jk}$ are now explicitly dependent on each macro-cell $k$ in the problem domain.  While the same feature functions $\varphi_j$ are employed for each $k$, the model can assign different coefficients  $\tilde{\theta}_{c, jk}$ at each $k$,  and therefore can potentially  account for local features in the coarse-graining process.
This increases the number of model parameters and in order to provide proper regularization as well as to enhance the interpretability of the results, we employ the same hyperparameters $\gamma_j$ for all $\tilde{\theta}_{c, jk}$ associated with the same feature function $j$, independently of the macro-cell $k$. 
In this manner information can be shared across macro-cells and feature functions will either be active or inactive over the whole domain. 
Predictive errors for the original and this enhanced model are compared in  Table \ref{nonLocalTable}   under  a low number of training  samples  $N = 16$ as well as for $N=1024$. 
 In the latter case, we observe that using different $\tilde{\theta}_{c,jk}$'s for different macro-cells $k$, leads to  improvements in predictive performance.
 For $N=16$ however, the simpler  model where $\tilde{\theta}_{c,jk} = \tilde{\theta}_{c,j}$ exhibits superior performance.

\section{Conclusions}
\label{sec:conclusions}
We have introduced a Bayesian formulation that performs simultaneous model-order and dimensionality reduction for problems characterized by high-dimensional  inputs/outputs as those arising in PDEs for random heterogeneous media. 
At the core of the proposed architecture  lies a coarsened version of the original description with a latent closure model (constitutive law). The  latter serves as a filter of the FOM high-dimensional input. The outputs of the coarsened model are  decoded in order to yield predictions of the FOM high-dimensional output. All three components are modeled with parametrized densities which are trained simultaneously using FOM simulation data. We have demonstrated that this can be achieved with only a few tens of such samples and that the resulting reduced-order model can extract essential information that allow it to produce crisp predictions even under different boundary conditions from those used in training.
The probabilistic nature of the model enables it to quantify uncertainties arising from the information loss that unavoidably takes place in all coarse-graining processes as well as those due to the use of finite-sized datasets. An essential feature of the model is the use of sparsity-inducing priors that promote the discovery of a low-dimensional set  of features of the  input which are most predictive of the FOM response. The training process involves Bayesian inference which is carried out using Stochastic Variational Inference tools that require repeated computations only of the coarse model and its  parametric derivatives.
Apart from uncertainty propagation, the resulting Bayesian reduced-order model can be readily used for other computationally intensive tasks such as optimization or the solution of inverse problems.

Several extensions can be envisaged with respect to all three building blocks. With regards to the coarse-graining density $p_c$ an important enhancement would involve the automatic discovery of the feature functions using semi-supervised models \cite{lawrence_semi-supervised_2004} rather than employing a predefined vocabulary. This would enable better predictive results as well as lead to further physical insight on the statistical descriptors of the underlying random medium that are predictive of its response. Several improvements are possible  for the coarse model employed. The immediate one is the development of an adaptive refinement scheme on the basis of probabilistic predictive metrics which would focus computational resources and statistical learning on the most informative  parts of the problem domain (i.e. subsets of the random input vector). The use of different physical models is also possible and  especially in multiscale problems, it might be necessary to employ a different description than the FOM.
Finally, with regards to the coarse-to-fine map $p_{cf}$, a possible enhancement could involve nonlinear maps between the coarse and FOM outputs that would promote further dimensionality reductions in this component.

\appendix

\section{Applied feature functions}
\label{sec:featureList}

\begin{table}[h]
\begin{center}
\tbf{Feature functions $\varphi$}
\vspace{2mm}

{\fontsize{3}{.7} \selectfont
\begin{tabular}{r|l|c} 
Index $j$ & Function $\varphi_j$ & Explanation \\
\hline
1 & constant & $\varphi_j = 1$ \\
\hline
2 & SCA & \makecell{$\varphi_j = \frac{\alpha + \sqrt{\alpha^2 + 4\lambda_{hi}\lambda_{lo}}}{2}$, $\alpha = \lambda_{lo}(2v_{lo} - 1) + \lambda_{hi}(2v_{hi} - 1)$} \\
\hline
3--4 & Maxwell-Garnett & $\varphi_j = \frac{\lambda_{\tm{mat}}}{1 - 2v_{\tm{inc}}}$ \\
\hline
5--6 & Differential Effective-Medium & $\left(\frac{\lambda_{\tm{inc}} - \varphi_j}{\lambda_{\tm{inc}} - \lambda_{\tm{mat}}}\right) \left(\frac{\lambda_{\tm{mat}}}{\varphi_j} \right)^{1/2} = 1 - v_{\tm{inc}}$\\
\hline
7--12 & Lineal path & \makecell{Lineal path function for certain phase/distance} \\
\hline
13--16 & Lin. path parameters & $a,b$ parameters of $a\cdot e^{-b\cdot d}$ fit to lineal path \\
\hline
17--18 & \multicolumn{2}{l}{Number of distinct high/low conducting blobs} \\
\hline
19--22 & \multicolumn{2}{l}{Number of high/low conducting pixels to cross from left to right/up to down} \\
\hline
23--26 & \multicolumn{2}{l}{Max. extent of high/low conducting blob in $x/y$--direction} \\
\hline
27--31 & Generalized mean & $\left(\frac{1}{M}\sum_{m = 1}^M (\lambda_{f,m}^{[k]})^q  \right)^{1/q}$ \\
\hline
32--37 & \multicolumn{2}{l}{Max./mean/variance of convex area of high/low conducting blobs} \\
\hline
38--41 & \multicolumn{2}{l}{Inv. distance of connected path through high/low cond. phase in $x/y$-direction, 0 if no connected path existent} \\
\hline
42--43 & Specific surface & $-4 \frac{\pa}{\pa d} \left. S_2(d)\right|_{d = 0}$, with 2-point correlation $S_2(r)$ \\
\hline
44--48 & ``Gaussian linear filter'' & \makecell{compute $ w_i = \mathcal N(\bs x_i| \bs \mu_{\tm{center}}, a\bs I)$ where $\bs \mu_{\tm{center}}$ is the macro-element center \\ and $x_i$ are fine-scale element locations. Compute $\varphi_j = \bs w^T \bs \lambda_f^{[k]}$} \\
\hline
49 & Standard deviation & $\varphi_j = \left<(\lambda_{f,i} - \left<\lambda_{f,i} \right>)^2 \right>$ \\
\hline
50 & Log standard deviation & $\varphi_j = \log(\left<(\lambda_{f,i} - \left<\lambda_{f,i} \right>)^2 \right>)$ \\
\hline
51 & Ising energy & Energy of a $2d$ Ising system with coupling $J = 1$ and no external field \\
\hline
52--63 & Two-point correlations & $\varphi_j = \frac{1}{N_{el, f}^[k]} \sum_{i = 1}^{N_{el,f}^{[k]}}\mathbb{1}_0(\lambda_{f, i}^{[k]} - \lambda_{f, i + d}^{[k]})$ \\
\hline
64--81 & Distance transformations & Mean/variance/maximum of distance transforms under different distance metrics \\
\hline
82--88 & Local PCA loadings & Perform PCA using every macro-cell $\bs \lambda_f^{[k]}$. Compute projections onto loadings $\varphi_j = \bs w^T \bs \lambda_f^{[k]}$ \\
\hline
89--92 & \multicolumn{2}{l}{Max. extent of high/low conducting blob in $x/y$--direction of whole microstructure $\bs \lambda_f$} \\
\hline
93--97 & \multicolumn{2}{l}{SCA, Maxwell-Garnett, Differential Effective Medium on whole microstructure $\bs \lambda_f$} \\
\hline
98--100 & Global PCA loadings & Perform PCA using whole microstructures $\bs \lambda_f$. Compute projections onto loadings $\varphi_j = \bs w^T \bs \lambda_f$
\end{tabular}
}
\vspace{2mm}
\end{center}
\caption{Set of 100 feature functions $\varphi$ applied in the $2d$ numerical examples.}
\label{featureTable}
\end{table}

Table \ref{featureTable} shows a list of the 100 feature functions used in the $2d$ numerical examples of Section \ref{2dexamples}. Features 1--88 take the subset $\bs \lambda_f^{[k]}$ as input, features 89--100 use the whole vector $\bs \lambda_f$.


\bibliographystyle{siam}
\bibliography{references_aug.bbl}
\end{document}